\documentclass{article}

\usepackage{arxiv}
\usepackage{booktabs}
\usepackage{algorithm}
\usepackage{algorithmic}
\usepackage[utf8]{inputenc} 
\usepackage[T1]{fontenc}    
\usepackage{hyperref}       
\usepackage{url}            
\usepackage{booktabs}       
\usepackage{longtable}
\usepackage{amsfonts}       
\usepackage{nicefrac}       
\usepackage{microtype}      
\usepackage{lipsum}		
\usepackage{graphicx}
\usepackage{makecell}
\usepackage{float}
\usepackage{amsmath}
\usepackage{enumitem} 
\usepackage{doi}
\usepackage{xcolor}
\usepackage{makecell}
\usepackage[style=apa, sorting=nyt, backend=biber]{biblatex}
\title{A Comprehensive Study of LLM-Based Argument Classification: from Llama through DeepSeek to GPT-5.2}

\author{{\hspace{1mm}Marcin Pietroń} \\
 	Faculty of Computer Science, \\ Electronics, and Telecommunications \\ AGH University of Krakow \\
 	\texttt{pietron@agh.edu.pl} \\
      \And
 {\hspace{1mm}Filip Gampel} \\
      Faculty of Humanities
 	 \\ AGH University of Krakow\\
 	\texttt{fgampel@agh.edu.pl} \\
  \And
      {\hspace{1mm}Jakub Gomułka} \\
 	Faculty of Humanities \\ AGH University of Krakow  \\
 	\texttt{jgomulka@agh.edu.pl} \\
    \And
      {\hspace{1mm}Andrzej Tomski} \\
 	Institute of Mathematics \\ University of Silesia  \\
 	\texttt{andrzej.tomski@us.edu.pl} \\
     	\And
        {\hspace{1mm}Rafał Olszowski} \\
 	Faculty of Humanities \\ AGH University of Krakow  \\
 	\texttt{rolszowski@agh.edu.pl} \\
 }

\date{}

\renewcommand{\headeright}{}
\renewcommand{\undertitle}{}
\renewcommand{\shorttitle}{A comprehensive study of LLM-based argument classification}

\begin{document}
\maketitle

\begin{abstract}

Argument mining (AM) is an interdisciplinary research field focused on the automatic identification and classification of argumentative components, such as claims and premises, and the relationships between them. Recent advances in large language models (LLMs) have significantly improved the performance of argument classification compared to traditional machine learning approaches. However, there remains a lack of systematic comparative evaluation of modern LLMs across multiple benchmark datasets, as well as a limited understanding of their error patterns and failure modes. This study presents a comprehensive evaluation of several state-of-the-art LLMs, including GPT-5.2, Llama 4, and DeepSeek R1, on large publicly available argument classification corpora such as Args.me and UKP. The evaluation incorporates advanced prompting strategies, including Chain-of-Thought prompting, prompt rephrasing, voting, and certainty-based classification. Both quantitative performance metrics and qualitative error analysis are conducted to assess model behavior.


The best-performing model in the study (GPT-5.2) achieves a classification accuracy of 78.0\% (UKP) and 91.9\% (Args.me). The use of prompt rephrasing, multi-prompt voting, and certainty estimation further improves classification performance and robustness. These techniques increase the accuracy and F1 metric of the modeuls by typically a few percentage points (from 2\% to 8\%). However, qualitative analysis reveals systematic failure modes shared across models, including instabilities with respect to prompt formulation, difficulties in detecting implicit criticism, interpreting complex argument structures, and aligning arguments with specific claims. The findings provide new insights into the strengths and limitations of LLMs in automated argument mining and highlight the importance of prompt engineering and ensemble techniques. This work contributes the first comprehensive evaluation that combines quantitative benchmarking and qualitative error analysis on multiple argument mining datasets using advanced LLM prompting strategies.



\end{abstract}


\section{Introduction}

Argument mining (AM) is an interdisciplinary research area spanning philosophy, logic, linguistics, rhetoric, law, psychology, and computer science. Within artificial intelligence, AM focuses on the automatic identification and analysis of argumentative structures in natural language texts. The goal is to extract components such as claims and premises, as well as the relationships between them, enabling structured representations of argumentative discourse \parencite{park-cardie-2018-corpus}. In recent years, AM has become an important subfield of Data Mining and Natural Language Processing (NLP), with applications in decision support systems, opinion analysis, legal informatics, and computational social science.

Early computational approaches to AM relied primarily on feature-based machine learning methods and manually engineered representations (\cite{cabrio2018five}, \cite{schaefer2022gercct}, \cite{lawrencereed}). With the rise of neural architectures, including recurrent and convolutional neural networks, performance improved substantially, particularly for component detection and relation classification tasks. The introduction of Transformer-based architectures, such as BERT, marked a significant breakthrough in NLP, enabling contextualized representations that improved performance in many AM benchmarks \parencite{pietron2024efficient, ref_stab18}. More recently, large language models (LLMs) such as Llama and GPT have demonstrated strong generalization capabilities across diverse reasoning and language understanding tasks. Despite these advances, the application of LLMs to structured argument classification remains insufficiently explored. Most standardized LLM benchmarks focus on multiple-choice reasoning, question answering, logical puzzles, or code generation. In contrast, argument classification requires the identification of implicit premises, pragmatic context, and nuanced semantic relations between propositions. Unlike many benchmark tasks with clearly defined rule-based solutions, argumentative reasoning often involves ambiguity, contextual interpretation, and subtle distinctions between support, attack, and neutrality relations. This makes argument classification a particularly challenging and informative testbed for evaluating the reasoning capabilities of LLMs.

Recent research has proposed prompting strategies that encourage intermediate reasoning steps, such as Chain-of-Thought prompting and tree-structured exploration mechanisms \parencite{wei2022chain}. These approaches aim to improve performance in tasks that require multi-step inference rather than direct answer prediction. However, it remains unclear whether such reasoning-oriented prompting strategies translate into measurable improvements in structured argument mining tasks. In particular, the extent to which LLMs can replicate or approximate human annotation behavior in argument relation classification is still an open question. In AM, the analytical pipeline is typically divided into three main subtasks of increasing complexity: (i) identification of argumentative components, distinguishing argumentative from non-argumentative segments; (ii) classification of component types, such as premises and claims; and (iii) identification of relational properties, determining whether one proposition supports, attacks, or is unrelated to another. The third subtask—argument relation classification—requires simultaneous semantic and logical analysis of multiple textual units and therefore constitutes a suitable framework to evaluate the reasoning abilities of modern LLMs.


This paper investigates the capability of contemporary LLMs to perform argument relation classification using two widely recognized datasets derived from established AM projects: UKP argument corpora and the Args.me corpus. Both datasets contain manually annotated argumentative relations, providing a reliable ground truth for systematic evaluation. We compare model predictions with human annotations and perform both quantitative and qualitative analyses to assess strengths and limitations. Our study demonstrates how different models, varying in size and reasoning capabilities, perform in argument classification tasks. The comparison is conducted using the Accuracy and F1-score metrics. The results reveal that the GPT model family outperforms the Llama and DeepSeek models. Additionally, we show that improving the prompt algorithm with rephrased prompts and incorporating a voting strategy can significantly improve performance, increasing Accuracy and F1 (from 2\% to 8\%).


The study is guided by the following research questions:

\textbf{RQ1. How does a particular syntax of prompts affect the quality of argument classification?}


\textbf{RQ2. What are the performance differences between different language models and how does the size of the LLMs increase the capabilities of argument classification?}

\textbf{RQ3. How do prompting and reasoning algorithms improve argument classification by LLMs?}

\textbf{RQ4. How can a certainty-based multiprompt improve the Accuracy of argument classification?}

\textbf{RQ5. What types of error do LLMs make in automatic argument classification?}

\textbf{RQ6. What are the shortcomings of existing annotated datasets commonly used to train or evaluate networks and models in argument classification, and how should they be improved?}

The main contributions of this work are:
\begin{itemize}

\item A systematic evaluation of state-of-the-art LLMs on two established argument mining datasets.

\item A comparative analysis of different prompting strategies.

\item A qualitative error analysis highlighting limitations of LLMs in structured argument reasoning.

\end{itemize}


To the best of our knowledge, this study presents one of the first evaluations across a wide spectrum of modern large language models (LLMs) on argument classification benchmarks. In addition, it introduces the development of an efficient prompt algorithm for analyzing arguments in natural language. Additionally, it presents a qualitative in-depth analysis of LLM errors in the argument classification process.

\section{Related works}
\label{sec:headings}

Argument classification (AC) is a specialized subtask in the broader field of AM. It concentrates on categorizing the identified elements of an argument into predetermined classes. These classes often include differentiating claims and premises, or identifying whether a component supports or opposes the argument. Additionally, argument classification can entail assessing the nature of the argument, such as determining its type or evaluating its quality (e.g., whether it is a strong or weak argument). This task is integral to constructing a structured representation of arguments. By clarifying the role and relationship of each component, it enhances our understanding of the dynamics within an argumentative discourse \parencite{lippi2016argumentation}. Argument classification (AC), as a specialized subtask within AM, was described by \textcite{daxenberger2020argumentext}, \textcite{lippi2016argumentation}, \textcite{dusmanu-etal-2017-argument} and many others. According to these authors, AC involves categorizing the identified components of an argument into predefined categories, such as differentiating between claims and premises or determining the stance of an argumentative component, whether supporting or opposing. 

Various data science techniques that utilize natural language processing have proven effective in AC and, more broadly, in AM. Initially, argument structures were often represented using trees or tree-like models, facilitating computation due to the availability of tree-based parsing techniques. However, real-world arguments frequently deviate from these idealized structures. More recently, researchers have shifted towards exploring non-tree-based argument structures in argument mining. Before the advent of BERT and other Transformer-based models, Support Vector Machines (SVMs) and neural networks were pivotal in AM, leveraging their pattern recognition and classification capabilities to identify and analyze argumentative structures in the text. 
Architectures such as Recurrent Neural Networks (RNN), e.g. Long Short-Term Memory (LSTM), and Convolutional Neural Networks (CNN) have been used to incorporate contextual information into machine decision-making processes. \textcite{niculae2017argument}  introduced the first non-tree model for argument mining, using a factor graph model combined with structured Support Vector Machines and Bidirectional Long Short-Term Memory. SVMs have been widely used in argument mining due to their effectiveness in binary classification tasks, which are well-suited to identifying whether a particular sentence or phrase is an argumentative component (e.g., claim vs. non-claim). Subsequently, \textcite{galassi2018argumentative} utilized LSTMs and residual network links to predict the connections among argument components. LSTM based approach for argument classification is also presented in \cite{ref_stab18}. These LSTM-based solutions have limits and can achieve up to 45\% accuracy in the UKP dataset, \parencite{ref_stab18, li2020empirical}.


While SVMs and neural networks significantly contributed to the development of AM, they had limitations, such as the need for extensive feature engineering and difficulties in capturing long-distance dependencies in text. The introduction of BERT (Bidirectional Encoder Representations from Transformers) and subsequent Transformer-based models revolutionized AM, \parencite{pietron2024efficient, li2020empirical}. The Transformer architecture, featuring its self-attention mechanism, was initially introduced by \textcite{vaswani2017} as a solution to the increasing computational and memory demands of recurrent neural networks (RNNs), which were considered state of the art at the time. 
Research on how transformers work, especially the BERT model \parencite{devlin-etal-2019-bert}, has attracted significant interest. Fine-tuning large pre-trained Transformer-based models resulted in remarkable performance gains across a wide range of tasks. Several recent studies have employed transformer-based models for argument mining. \textcite{reimers2019classification} leveraged contextual word embeddings such as BERT and ELMo to significantly enhance argument/non-argument classification and proposed methods for argument clustering. \textcite{chakrabarty2019ampersand} introduced a BERT-based model for argument component classification and relation detection within persuasive online conversations. Going further, \textcite{chen2021bert} is using pre-trained BERT based models for predicting arguments where the structure forms a directed acyclic graph. Moreover, \textcite{ruiz2020transformer} present an analysis of the behavior of transformer-based models (i.e., BERT, XLNET, RoBERTa, DistilBERT and ALBERT) when predicting argument relations, and evaluate the models in five different domain specific corpora, with the objective of finding the less domain dependent model. The work presented in \cite{li2020empirical} shows that DistilBERT achieves 54.2\% and the BERT precision is around 57.7\% in the UKP benchmark. In \cite{pietron2024efficient}, the authors present a hybrid of BERT and ChatGPT-4. The proposed architecture significantly outperforms other ML-based solutions. It achieves 89.5\% accuracy in the Args.me and 68.5\% on the UKP benchmark. In this work, the LLM is used as an additional classifier on selected subsets of arguments for which the BERT model shows low response confidence. The main shortcomings of this work are the lack of evaluation of the GPT-4 model on entire datasets and the lack of in-depth error analysis. Recent Transformer-based models include BLOOM (\cite{lescao2023}), Llama (\cite{touvron2023}), GPT-5 and DeepSeek-R1 (\cite{guo2025}). There is a lack of works which study the performance of these models in argument mining, especially those with reasoning capabilities. Chain-of-Thought is an approach that simulates human-like reasoning processes by delineating complex tasks into a sequence of logical steps towards a final solution. This methodology reflects a fundamental aspect of human intelligence, offering a structured mechanism for problem-solving. This technique can be applied in prompt or incorporated into the LLM, forming a "reasoning model".

Given the challenges posed by fully automated AM methods, an interesting research direction has emerged in exploring hybrid approaches that combine the efforts of human annotators with AI. 
The automated AM methods often struggle to determine whether two arguments express the same viewpoint (\cite{chakrabarty2019ampersand}; \cite{daxenberger2017essence}), and reliance on a limited set of labeled data can lead to the exclusion of minority opinions, thereby creating a bias towards more popular or frequently repeated arguments. For example, a hybrid method called HyEnA \parencite{vandemeer2024hybrid} employs a sampling algorithm that guides human annotators individually through an opinion corpus. Then, an intelligent merging strategy helps annotators combine their results into clusters of arguments, integrating both manual and automatic labeling.

\section{Data}

\paragraph{Dataset choice} In our research, we decided to conduct a comparative study using two corpora containing different argument datasets, each of which was developed by a different research team. These are: the UKP corpus \parencite{Stab2018} and the Args.me corpus \parencite{ajjour2019data}. These corpora have gained recognition in recent argument mining research, but they have not yet been studied comparatively or with recent large language models such as Llama, DeepSeek or GPT-5. Previous studies, including \cite{bar-haim-etal-2017-stance}, \cite{boltuzic2014back},  \cite{Stab2018} were limited to research on a single corpus, and mostly covered pre-transformer-based NLP technologies.

Both corpora used in our study share fundamental structural characteristics of argumentation, namely:

\begin{itemize}
    \item Presence of a thesis – a central conclusion or topic around which the argumentation is constructed; 
    \item  Provision of a list of premises – each labeled as supporting or opposing the thesis, or with no relation (UKP)
\end{itemize}

The form and quality of argument descriptions vary between datasets; for instance, some include off-topic entries, non-arguments, or rephrased conclusions in place of genuine premises.

The UKP corpus \parencite{Stab2018} comprises datasets of arguments derived from online comments on eight controversial topics: abortion, cloning, the death penalty, gun control, minimum wage, nuclear energy, school uniforms, and marijuana legalization. This corpus includes over 25,000 instances. The sentences were independently annotated by seven individuals recruited through the Amazon Mechanical Turk (AMT) crowdsourcing platform. For each classification, an agreement level was required, with Cohen’s kappa ($\kappa$) set at 0.723, surpassing the commonly accepted threshold of 0.7 for reliable results \parencite{carletta1996assessing}. The classification labels used were: (1) supporting argument (Argument\_for), (2) opposing argument (Argument\_against), and (3) non-argument (NoArgument).

The Args.me corpus (version 1.0, cleaned) provided by \textcite{ajjour2019data} consists of arguments collected from four debate portals in mid-2019: Debatewise, IDebate.org, Debatepedia, and Debate.org. The arguments were extracted using heuristics specifically designed for each debate portal. The datasets used in our simulations are Idebate.org, Debatepedia, and Debatewise, which together contain 47,992 arguments. Debatepedia has the highest number of arguments, while IDebate.org shows the least disproportion between PRO and CON arguments, with the greatest class imbalance observed in Debatepedia. The annotations in this dataset include conclusions and premises, which are further categorized into PRO premises (arguments supporting the thesis) and CON premises (arguments opposing the thesis).

\paragraph{Samples used in this study}
We performed an initial screening of all datasets to crop very long records (>2000 characters in the \texttt{argument}/\texttt{sentence} field). In the context of this study, these will be referred to as the "full" sets. For each dataset, we then created a "trimmed" version of 2000 records for heavy calculations, involving large or proprietary models, etc. The trimmed versions were sampled randomly from the full sets, ensuring that the original class imbalance is preserved. The record counts of both sets can be found in tables \ref{tab:UKP_classcount} and \ref{tab:argsme_classcount}. Appendix \ref{appendix:results} shows detailed sample sizes for each calculation performed in the study.

\begin{table}[ht]
\centering
\caption{UKP class counts for full and trimmed datasets}
\small
\begin{tabular}{lrrrrrrrr}
\toprule
\textbf{Dataset} & \textbf{Full For} & \textbf{Full Against} & \textbf{Full NoArg} & \textbf{Full Total} & \textbf{Trim For} & \textbf{Trim Against} & \textbf{Trim NoArg} & \textbf{Trim Total} \\
\midrule
abortion & 680 & 822 & 2427 & 3929 & 346 & 418 & 1236 & 2000 \\
cloning & 706 & 839 & 1494 & 3039 & 465 & 552 & 983 & 2000 \\
death & 457 & 1111 & 2083 & 3651 & 250 & 609 & 1141 & 2000 \\
gun & 787 & 665 & 1889 & 3341 & 471 & 398 & 1131 & 2000 \\
marijuana & 587 & 626 & 1262 & 2475 & 474 & 506 & 1020 & 2000 \\
nuclear & 606 & 852 & 2118 & 3576 & 339 & 476 & 1185 & 2000 \\
school & 545 & 729 & 1734 & 3008 & 362 & 485 & 1153 & 2000 \\
wage & 576 & 551 & 1346 & 2473 & 466 & 446 & 1088 & 2000 \\
\bottomrule
\end{tabular}

\label{tab:UKP_classcount}
\end{table}

\begin{table}[ht]
\centering
\caption{Args.me class counts for full and trimmed datasets}
\small
\begin{tabular}{lrrrrrr}
\toprule
\textbf{Dataset} & \textbf{Full For} & \textbf{Full Against} & \textbf{Full Total} & \textbf{Trim For} & \textbf{Trim Against} & \textbf{Trim Total} \\
\midrule
debatepedia & 15787 & 5406 & 21193 & 1490 & 510 & 2000 \\
debatewise & 8109 & 5642 & 13751 & 1179 & 821 & 2000 \\
idebate & 6445 & 6603 & 13048 & 988 & 1012 & 2000 \\
\bottomrule
\end{tabular}

\label{tab:argsme_classcount}
\end{table}




\section{Experimental setup}

\subsection{System architecture and model choice}

\begin{figure}[h]
    \centering
    \includegraphics[width=15cm]{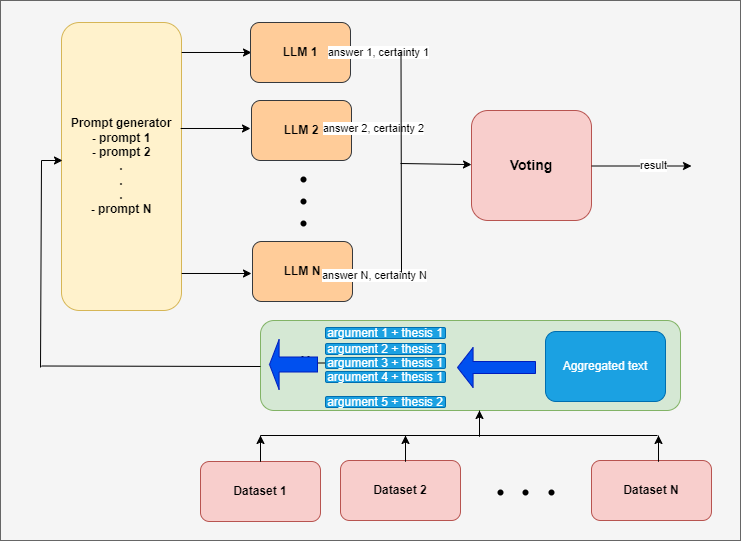}
    \caption{System architecture with voting strategy}
    \label{fig:architecture}
\end{figure}

The general architecture of the proposed approach is illustrated in Fig. \ref{fig:architecture}. The input to the system comes from the different datasets. From the aggregated data, arguments are forwarded to the Prompt Generator module; optionally, a thesis statement may be included.

The Prompt Generator produces a set of prompts instructing the models to classify the relation of the argument to the thesis, and to estimate the certainty of their answer. If the thesis is not provided (as in the case of the UKP corpus), the Prompt Generator injects a thesis based on the argument domain. In the context of this study, for each argument-thesis pair we generate $N=4$ prompts.

Subsequently, a set of instances of a LLM responds to the generated prompts. The output of each instance is parsed into a label prediction along with a confidence score. For our ensemble method, the outputs are then passed to the Voting module, which applies several ensemble voting strategies to generate the final system prediction. As we will show, the voting mechanisms help mitigate errors produced by individual prompts. 

In our study, we test the performance of various recent popular large language models, coming from the Llama, Deepseek and GPT family. Details on all models chosen for the study are provided in Appendix \ref{appendix:models}.

\subsection{Prompting strategies}
\label{subsec:promptin_strategies}

Our study focuses on assessing the performance of general-purpose LLM's without any prior specific training on a similar task or dataset related to argument mining. All experiments are related to the task of argument classification. Since prompting techniques severely impact the performance of LLM models in many areas, an attempt was made to examine some of them.  

\paragraph{mRAR prompting} Rephrase and Respond (RaR) is a prompting technique in which LLM's are asked to rephrase and expand questions posed by the user and provide responses to these \parencite{deng2023rephrase}. For the purpose of our study, we develop a standard set of four prompts, all aimed at the same task of classifying the stance of an argumantative text with respect to a topic or thesis. We will call this manual rephrased and respond strategy (mRAR). The prompts differ in:

\begin{itemize}
    \item \textbf{Response format}: Two prompts ask the model to format the response as a single letter (F/A/N), while two ask for verbal labels (For/Against/No argument).
    \item \textbf{Thesis presentation}: Two prompts are rather verbose, while the other two feature a shorter and simpler formulation of the task.
\end{itemize}

\begin{figure}[h]
    \centering
    \includegraphics[width=8cm]{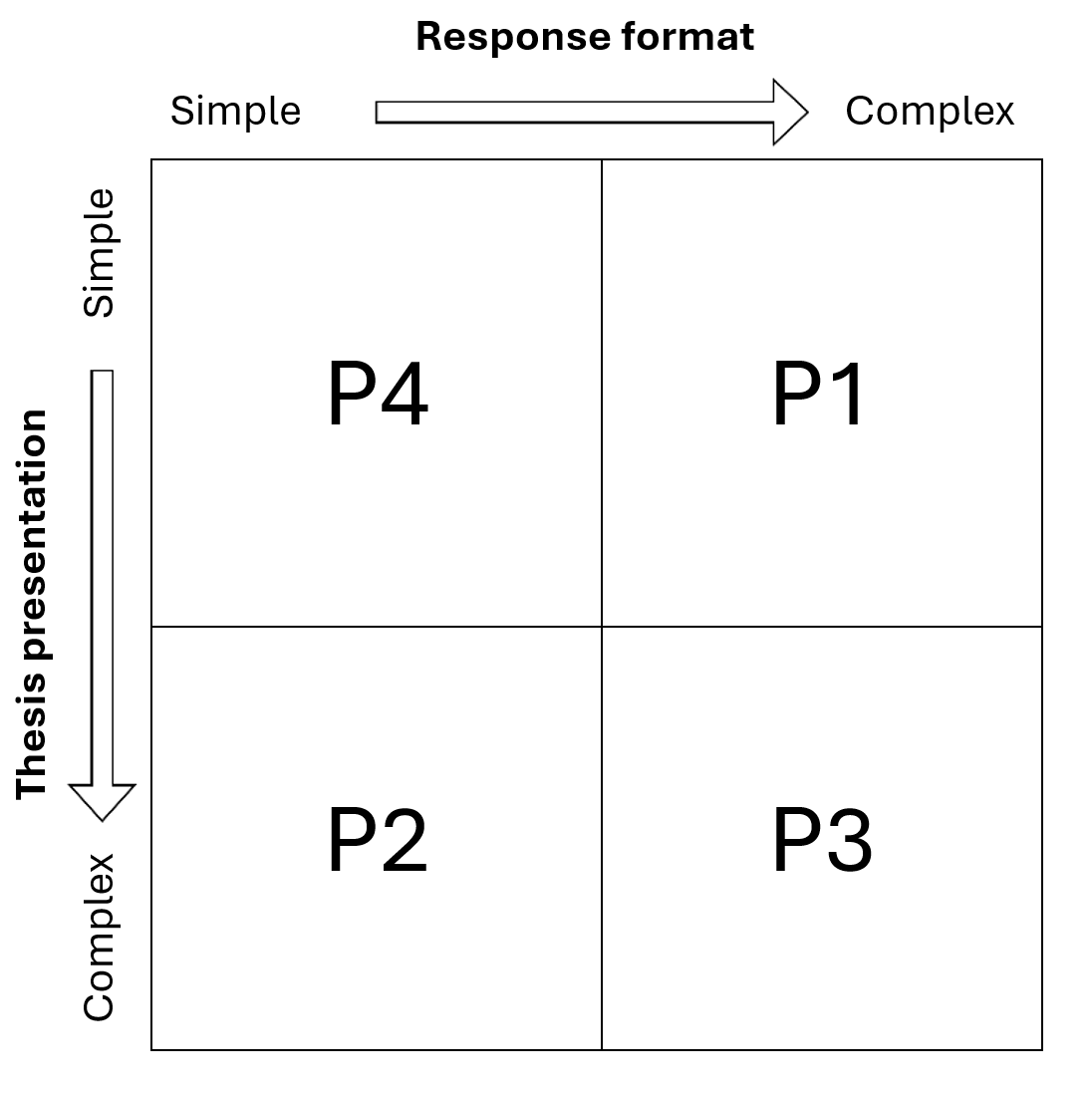}
    \caption{Overview of prompt complexity used in the study.}
    \label{fig:prompts}
\end{figure}

These relations are visualized in Fig. \ref{fig:prompts}. The prompts were minimally adapted to match the dataset format (three answer options for UKP vs. two for Args.me). The full texts of all prompts can be found in Appendix \ref{appendix:prompts}.

\paragraph{Chain-of-Thought prompting} The Chain of Thought (CoT) technique aims at mirroring human reasoning and problem-solving through a coherent series of logical deductions \parencite{wei2022chain}. As a prompting strategy, the LLM is usually asked to break down the task into smaller steps, sequentially analyzing and summarizing all available information and context before answering \parencite{zhou2023thread}. At the time of writing, "Reasoning models" \parencite{besta2025} are the de-facto industry standard for proprietary and high-end language models. These models are specifically trained to perform multiple steps of logical reasoning in order to solve complex tasks. However, some popular open-weight models like the Llama series do not have inbuilt reasoning capacities. For these models, we study the effect of explicitly encouraging the model to arrive at its answer through step-by-step reasoning (see Appendix \ref{appendix:reasoning_prompt}).

\paragraph{Certainty self-assessment} Finally, for the four mRaR prompts, a second prompt was appended to the conversation, after the model provided its classification answer. The prompt asked the model to assess how certain its answer was on a percentage scale from 0 to 100 (see Appendix \ref{appendix:certainty_prompt}). These self-assessments allow as to combine the answers from the four prompts through a number of voting algorithms to obtain a final aggregate answer. The algorithms are described in detail in the following section.

\subsection{Voting algorithms}
\label{subsec:voting_algorithms}
The classification of the text $p$ using the prompt $i$ by the model $T_\Theta$, together with a certainty rating, may be formally denoted as

\begin{equation}
T_{\Theta}(p_{i}) \mapsto \mathcal{L} \times [0,1]
\label{eq:answer}
\end{equation}

Here $\mathcal{L} =\{F, A, N\}$ is the set of possible predictions, where $F$ indicates an argument predicted as \textbf{'For'}, $A$ the prediction \textbf{'Against'}, and $N$ signifies \textbf{'No Argument'}. Let $T_{\Theta}(p_i) = (q_i, c_i)$ denote the prediction of a label $q_i \in \mathcal{L}$ with certainty $c_i \in [0,1]$.

We will consider three versions of the aggregated voting alogrithm:
\begin{itemize}
\item \textbf{Simple vote} The most frequent answer among the promptings is chosen. In case of a tie, the result is chosen randomly from the winning options. This version of the voting algorithm ignores the certainty self-ratings.

Let $m$ be the number of prompts. For each class $q \in \{F, A, N\}$, define the vote count:

\begin{equation}
    V(q) = \sum_{i=1}^{m} \mathbf{1}\!\left[q_i = q\right],
\end{equation}
where $\mathbf{1}[\cdot]$ denotes the indicator function. Now define the set of winners:
\begin{equation}
\mathcal{W} = \left\{ q \in \mathcal{L} \;:\; V(q) =  \max_{k \in \mathcal{L}} V(k) \right\}.
\end{equation}
The final classification is given by ($\mathcal{U}$ is the uniform distribution):
\begin{equation}
    q_{pred} \sim \mathcal{U}(\mathcal{W})
\end{equation}

\item \textbf{Tiebreak vote} Like in simple vote, the result is the most frequent answer. However in case of a tie, the certainty self-ratings are used to determine the winner.

For $\mathcal{W}$ as above, define the aggregated certainty score:
\begin{equation}
    S(q) = \sum_{i=1}^m c_i \,\mathbf{1}\!\left[q_i = q\right], \quad q \in \mathcal{W}
\end{equation}
Then the final prediction is:
\begin{equation}
    q_{pred} = \arg\max_{q \in \mathcal{W}} S(q)
\end{equation}
\item \textbf{Weighted vote} For each category, a certainty-weighted sum of votes is calculated. The winner is the option with the highest score.

For all classes $q$, consider certainty score:
\begin{equation}
    S(q)=\sum_{i=1}^{m}c_i \,\mathbf{1}\!\left[q_i = q\right].
\end{equation}
The predicted label is:
\begin{equation}
    q_{pred}= \arg\max_{q \in \mathcal{L}} S(q).
\end{equation}
\end{itemize}
The Tiebreak vote is presented in Algorithm \ref{algorithm}. The main loop iterates over the set of prompts and queries the model for the argument classification and the certainty of its answer (lines from 3 to 15). After each answer, it updates the aggregated results.
Then it checks which classes were recognized most frequently (line 16). If there is only one such class, it constitutes the final answer (line 19). Otherwise, the confidence level (certainty) of the answers is taken into account and the final predicted class is determined on the basis of this (line 21).

\begin{algorithm}
\begin{algorithmic}[1]
\REQUIRE{$L$ -- LLM model returning $(q,c)$}
\REQUIRE{$\Omega$ -- list of prompts}

\STATE{$\delta_f, \delta_a, \delta_n \gets 0$}
\STATE{$c_f, c_a, c_n \gets 0$}

\FOR{$\omega \in \Omega$}
    \STATE{$(q,c) \gets L(\omega)$}
    
    \IF{$q ==$ \textbf{'For'}}
        \STATE{$\delta_f \gets \delta_f + 1$}
        \STATE{$c_f \gets c_f + c$}
    \ELSIF{$q ==$ \textbf{'Against'}}
        \STATE{$\delta_a \gets \delta_a + 1$}
        \STATE{$c_a \gets c_a + c$}
    \ELSE
        \STATE{$\delta_n \gets \delta_n + 1$}
        \STATE{$c_n \gets c_n + c$}
    \ENDIF
\ENDFOR

\STATE{$\Delta \gets \max(\delta_f,\delta_a,\delta_n)$}
\STATE{$\mathcal{W} \gets \{k \in \{f,a,n\} : \delta_k = \Delta\}$}

\IF{$|\mathcal{W}| = 1$}
    \STATE{$out \gets k \text{ such that } \mathcal{W} = \{k\}$}
\ELSE
    \STATE{$out \gets \arg\max_{k \in \mathcal{W}} c_k$}
\ENDIF

\RETURN{$out$}

\caption{Majority Voting with Certainty-Based Tie-Breaking}
\label{algorithm}
\end{algorithmic}
\end{algorithm}

\section{Numerical results}

In this section results from running a wide range of argument classification experiments are reported. To better assess the reliability of the final model performance calculations, we will first analyze how model performance depends on factors like prompt formulation, dataset and sampling randomness. 

\subsection{Variability}
\paragraph{Prompt variability (RQ1)}
\label{subsec:prompt_variability}

\begin{figure}[h]
\centering
\includegraphics[width=14 cm]{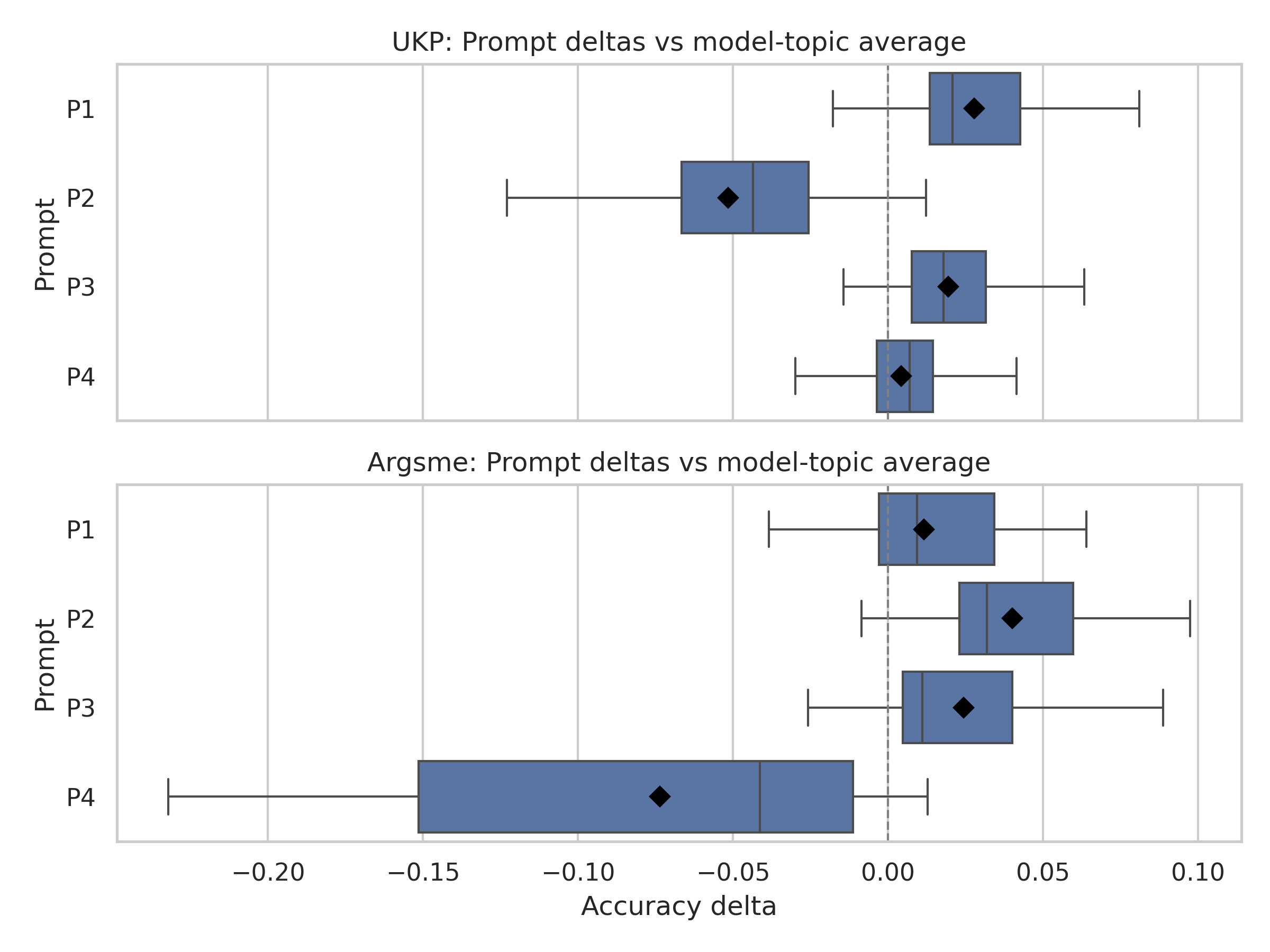}
\caption{Impact on accuracy due to prompt choice (calculations exclude Llama 3.1 1B)}
\label{fig:prompt-boxplots}
\end{figure}

As described in Sec. \ref{subsec:promptin_strategies}, the two main distinctions within our basic prompt set are answer format (P1,P3 vs. P2,P4) and verbosity (P1,P4 vs P2,P3). The effect of prompt choice with respect to average performance is shown in Fig. \ref{fig:prompt-boxplots}. First we calculate the accuracy for each model-dataset pair, averaged over all four prompts. Then we calculate the deviations of the accuracy for the particular prompts. The distribution of these deviations for all model-prompt-dataset combinations, grouped by prompt and corpus, is shown in the figure. 

We first note that, on average, classification with natural language answer format (P1,P3) works better than the abbreviated symbolic format (P2,P4). This is especially clear in the case of the UKP corpus. Furthermore, more elaborate formulation of the task (P2,P3) seem preferable to the shorter formulation (P1,P4). This certainly holds for the Args.me database. In this corpus, we note a marked instability for P4, which combined symbolic format with a brief task formulation. For the UKP corpus, the effect of prompt verbosity is not as clear: The verbose prompt P2 shows notably poor performance. This may be related to the different data structures between the corpora: Within Args.me, each record provides a pair of thesis and argument. In contrast, the UKP corpus consist only of arguments, grouped into datasets by discussion topic. The thesis is taken to be implied, since these topics are usually highly controversial issues, dividing debate participants clearly into two opposing camps. For the purpose of prompt 2, we attempted to reconstruct the theses which would best characterize these opposing standpoints (see Appendix \ref{appendix:prompts}). In cases where the scope of the original debate was broader than the reconstructed thesis, this might have impaired the argument classification. Thus for UKP, higher accuracy is achieved when stating the thesis very roughly (e.g. being simply "for" or "against" abortion, minimum wage etc.)

\paragraph{Variance between datasets}

\begin{figure}[h]
\centering
\includegraphics[width=14 cm]{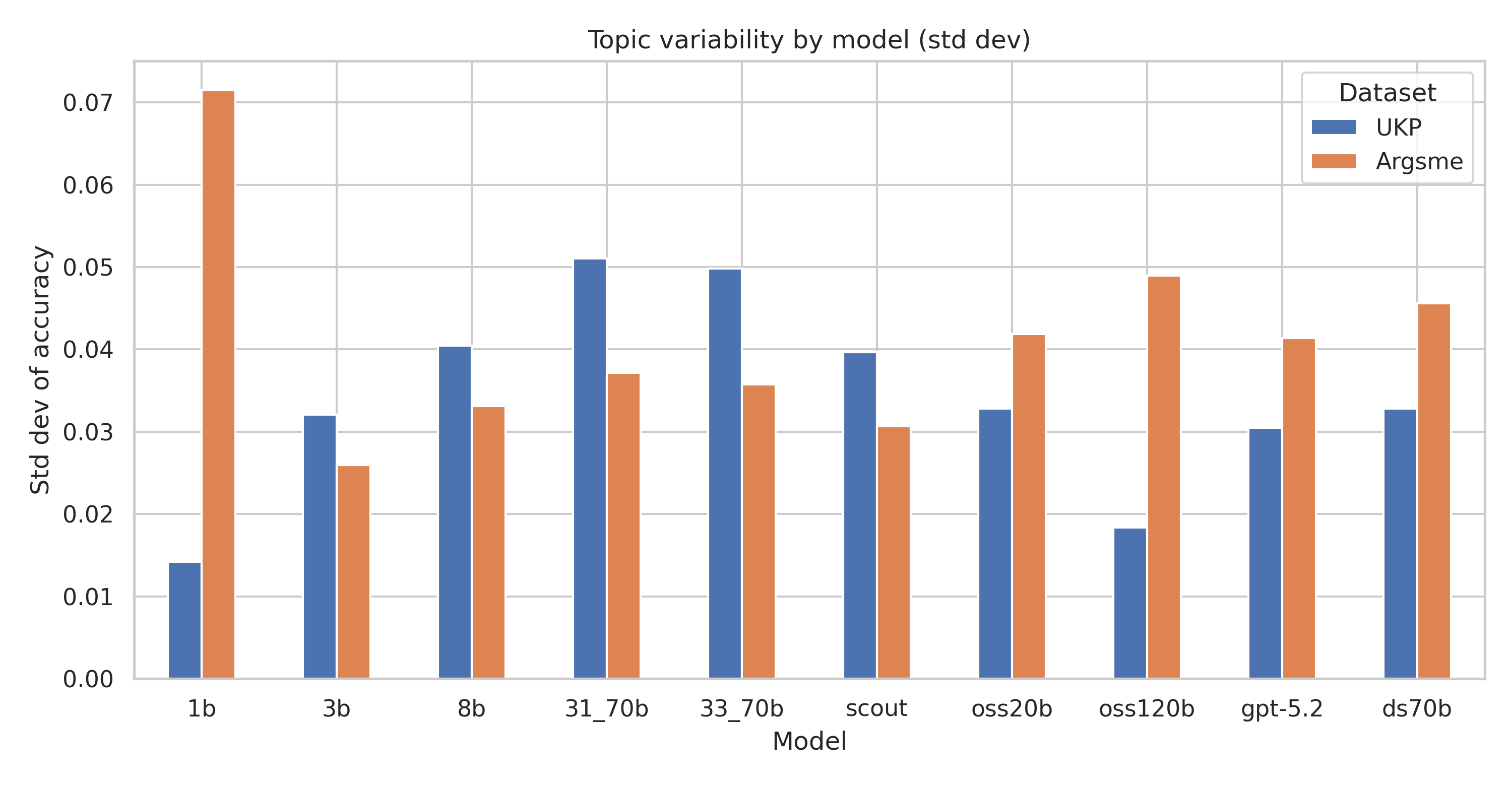}
\caption{Standard deviation from model average depending on dataset}
\label{fig:dataset_variance}
\end{figure}

We now calculate how model performance varies with different datasets. For each model, we take its average accuracy, and calculate the standard deviation based on performance in model-dataset pairs. We again perform calculations seperately on UKP and Args.me data. The Llama models have a notably higher variance for the UKP datasets, while for Args.me, GPT and Deepseek performance varies slightly more. All in all however, for a given dataset the typical deviation from average is around 2 to 4 percentage points. We also note that some datasets lead to systematically poorer classification accuracy then others (cf. Fig. \ref{heatmap1}). This is particularly true for the debatewise dataset within Args.me. One obvious reason is variance in data and annotation quality. However, the differences could also stem from LLM bias, as discussed in Sec. \ref{sec:domain-specific errors}.

\paragraph{Sampling variance due to temperature}
\begin{figure}[h]
\centering
\includegraphics[width=14 cm]{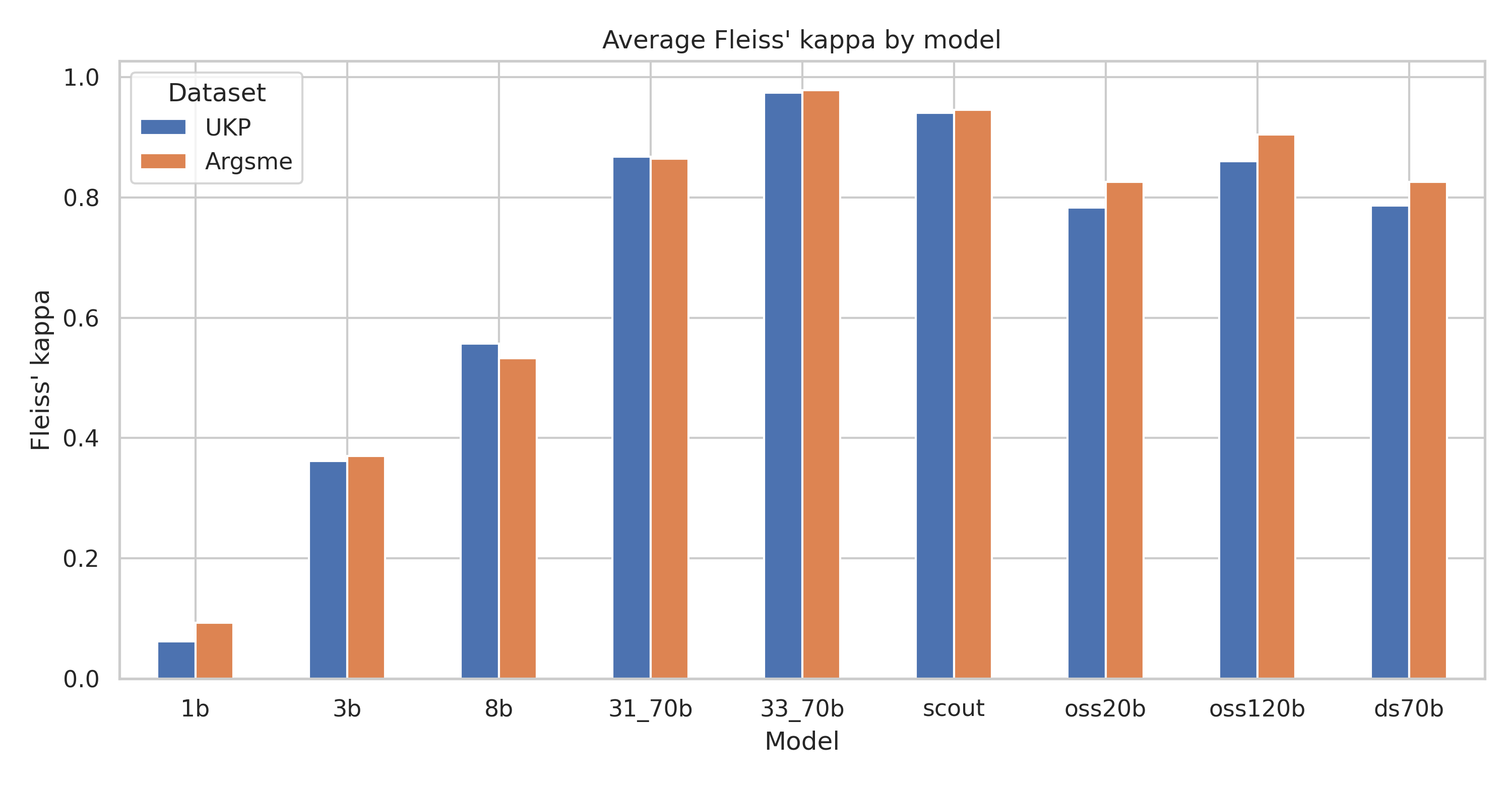}
\caption{Average Fleiss' kappa by model for UKP and Args.me.}
\label{fig:variability-kappa}
\end{figure}

In our study, we consequently set the temperature to 0.6, which is the recommended temperature for the Llama models. This introduces randomness into the LLM output. To estimate how strongly the classification varies due to this decoding strategy, we perform a variability study: We use the "trimmed" datasets to repeatedly perform inference on identical prompts. Due to cost restraints, the study is performed on models which we were able to run locally (i.e. all except GPT 5.2). For each record, prompt and model inference is performed three times. We then calculate the Fleiss kappa, which is a measure of agreement for raters in categorical ratings \parencite{Fleiss1971}. It is a generalization of the popular Cohen kappa, which measures agreement between two raters.

We note that intra-model agreement is especially strong for large, well-performing models. This is an intuitive result assuming that stronger models more often "know" the answer and less frequently resort to "guessing", as is the case with smaller models. The typical $\kappa$ for models larger than Llama 8b is above 0.8, which indicates very good agreement for these models.

\subsection{Model performance}
We now turn to assess how the models in our study compare to each other in argument classification. For the accuracies in this section, we always report the standard error. Unless stated otherwise, for each model we first average accuracy across prompts within each dataset. We then compute the mean and standard error across datasets to obtain the values reported in the tables and figures.

\paragraph{Zero-shot accuracy (RQ2)}

\begin{figure}[h]
\centering
\includegraphics[width=14 cm]{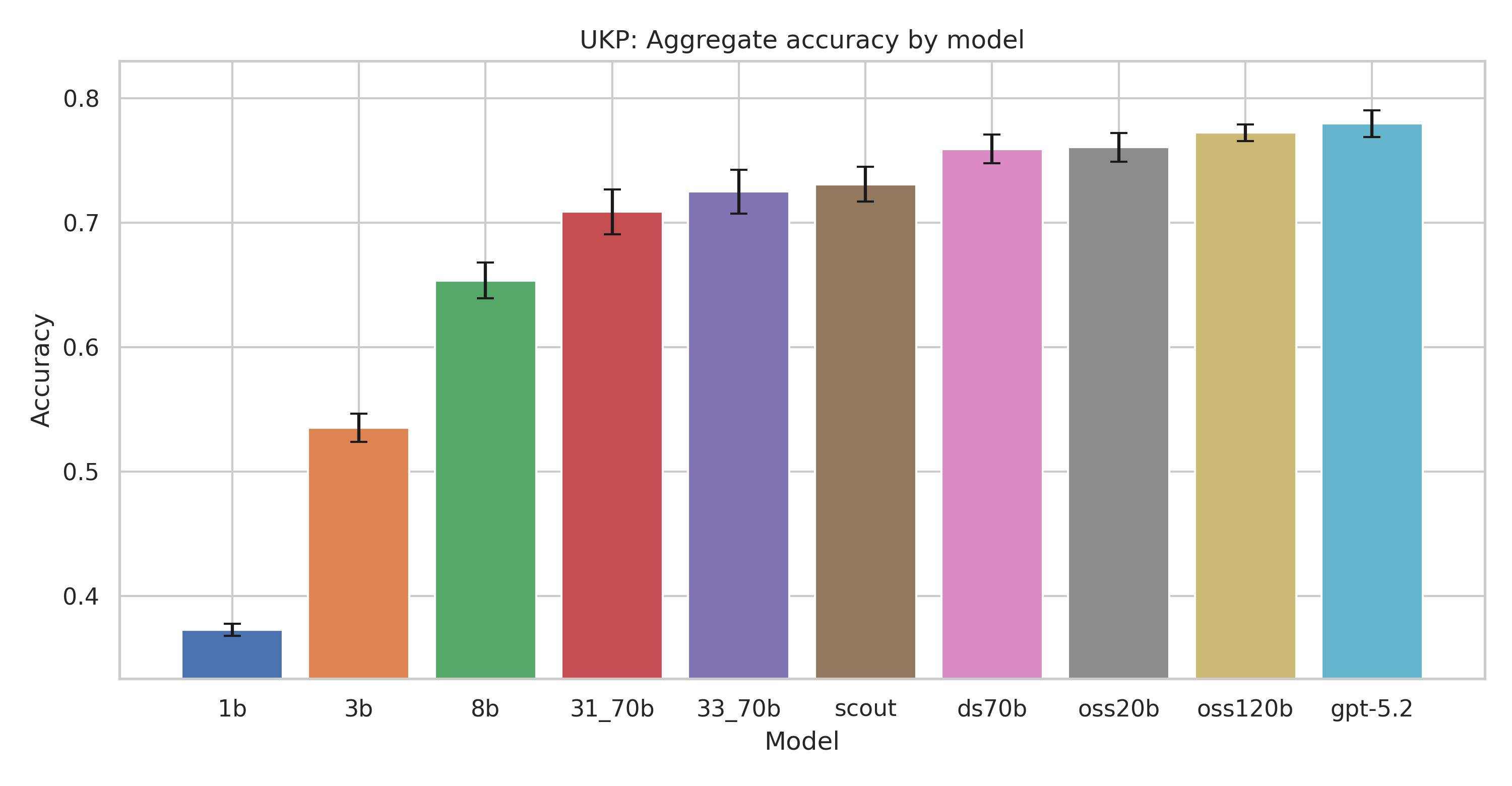}
\caption{UKP aggregate Accuracy by model}
\label{fig:aggregate-ukp}
\end{figure}

\begin{figure}[h]
\centering
\includegraphics[width=14 cm]{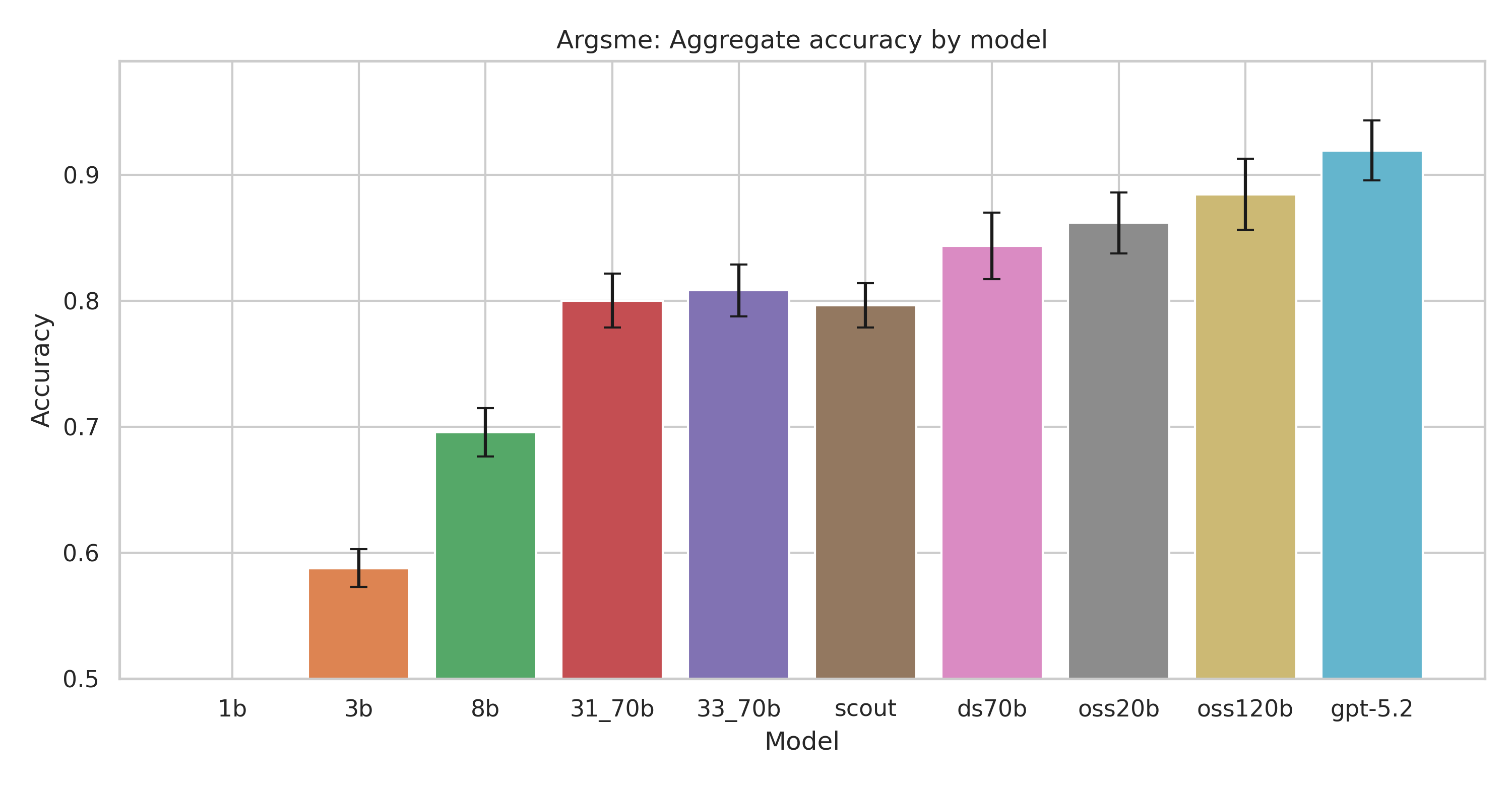}
\caption{Args.me aggregate Accuracy by model}
\label{fig:aggregate-argsme}
\end{figure}

We begin by comparing the accuracy of the models in the absence of prompting strategies. Figures \ref{fig:aggregate-ukp} and \ref{fig:aggregate-argsme} illustrate the performance averaged across the four basic prompts and all datasets, relative to the random guessing baselines (33.3\% for UKP and 50\% for Args.me). As anticipated, performance declines significantly for the smallest models. While Llama 8B and 3B may still be utilized under severe computational constraints, the 1B model’s performance is comparable to random guessing, rendering it ineffective for the task in a zero-shot capacity. Consequently, we exclude it from most subsequent analyses. (Note: For the Args.me corpus, Llama 1B’s performance falls below the 50\%). 

Only minuscule performance differences are observed between Llama 3.1 70b, 3.3 70b and Llama 4 Scout. As expected, the flagship OpenAI model GPT-5.2 -- the only proprietary model in the study -- shows the best performance. Second ranking is the large GPT oss-120b. ds70b,  which is the Llama 3.3 70b model fine tuned with Deepseek R1, shows notable improvement over its base model. Perhaps most noteworthy is the performance of GPT oss-20b. With a modest parameter count, it significantly outperforms the much larger Llama 70b models. The authors believe this speaks to the potential of achieving argument mining performance on par with state-of-the-art general purpose models using aptly tuned, much smaller open-weight models like GPT oss-20b. Investigating this possibility represents a promising area for future research stemming from this work.

\paragraph{Chain-of-thought prompting (RQ3)}

\begin{figure}[h]
\centering
\includegraphics[width=16 cm]{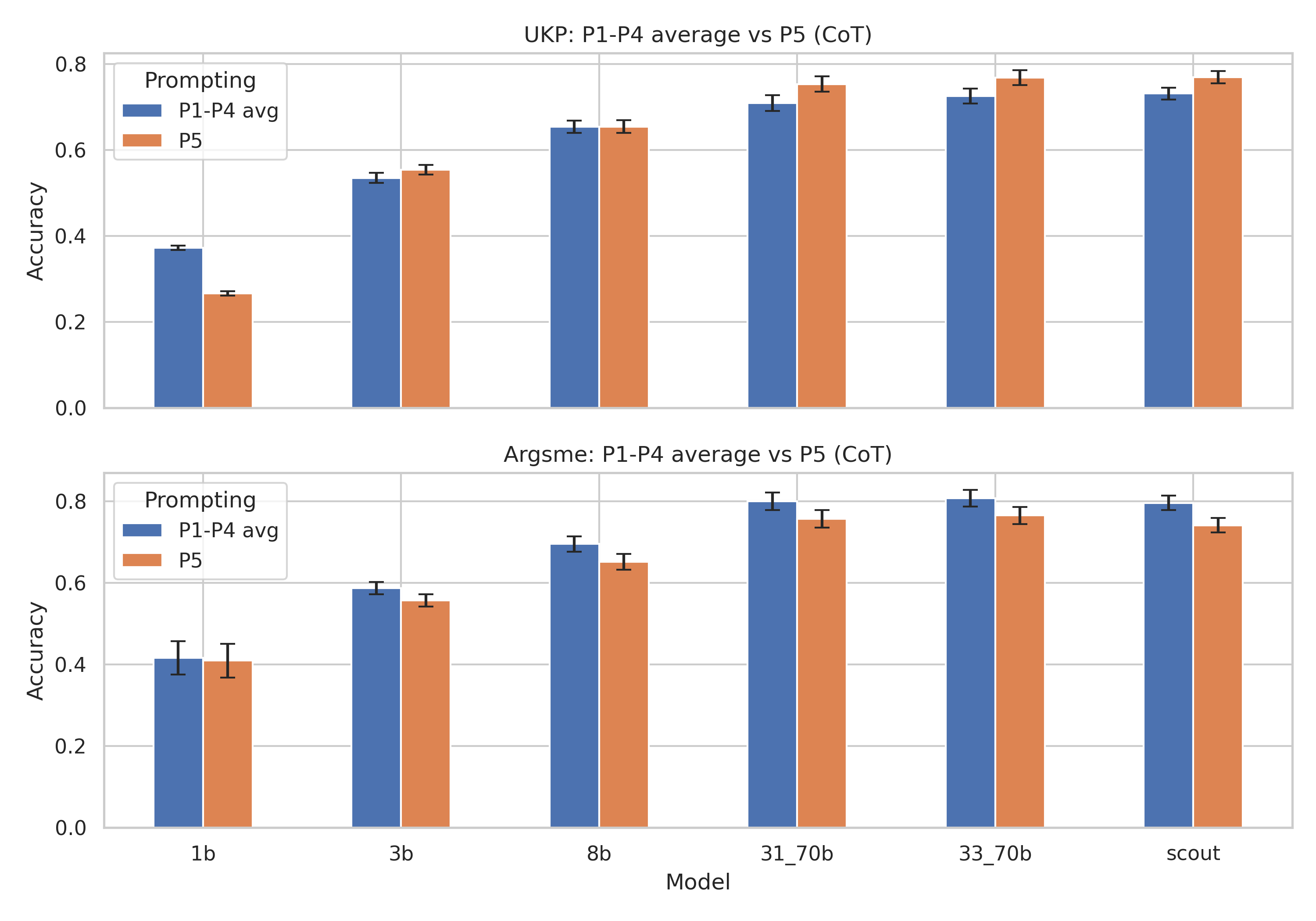}
\caption{P1--P4 average vs P5 (CoT) accuracy for UKP and Args.me.}
\label{fig:cot-prompting}
\end{figure}

As mentioned, explicit chain-of-thought prompting was performed only for the Llama models, since these are the only models in the studies which do not have an inbuilt "reasoning mode". The models were encouraged to perform step-by-step reasoning using a special prompt format, see Appendix \ref{appendix:reasoning_prompt}. The resulting Accuracy is compared to average zero-shot performance in Fig. \ref{fig:cot-prompting}. The efficacy of the CoT prompting method in this context is rather mixed. Improvements of a few percentage points were noted for the larger models as well as Llama 3b in case of the UKP corpus. On the other hand, performance worsened for all models in case of the Args.me corpus.

\paragraph{Voting algorithms (RQ4)}
\begin{figure}[h]
\centering
\includegraphics[width=14 cm]{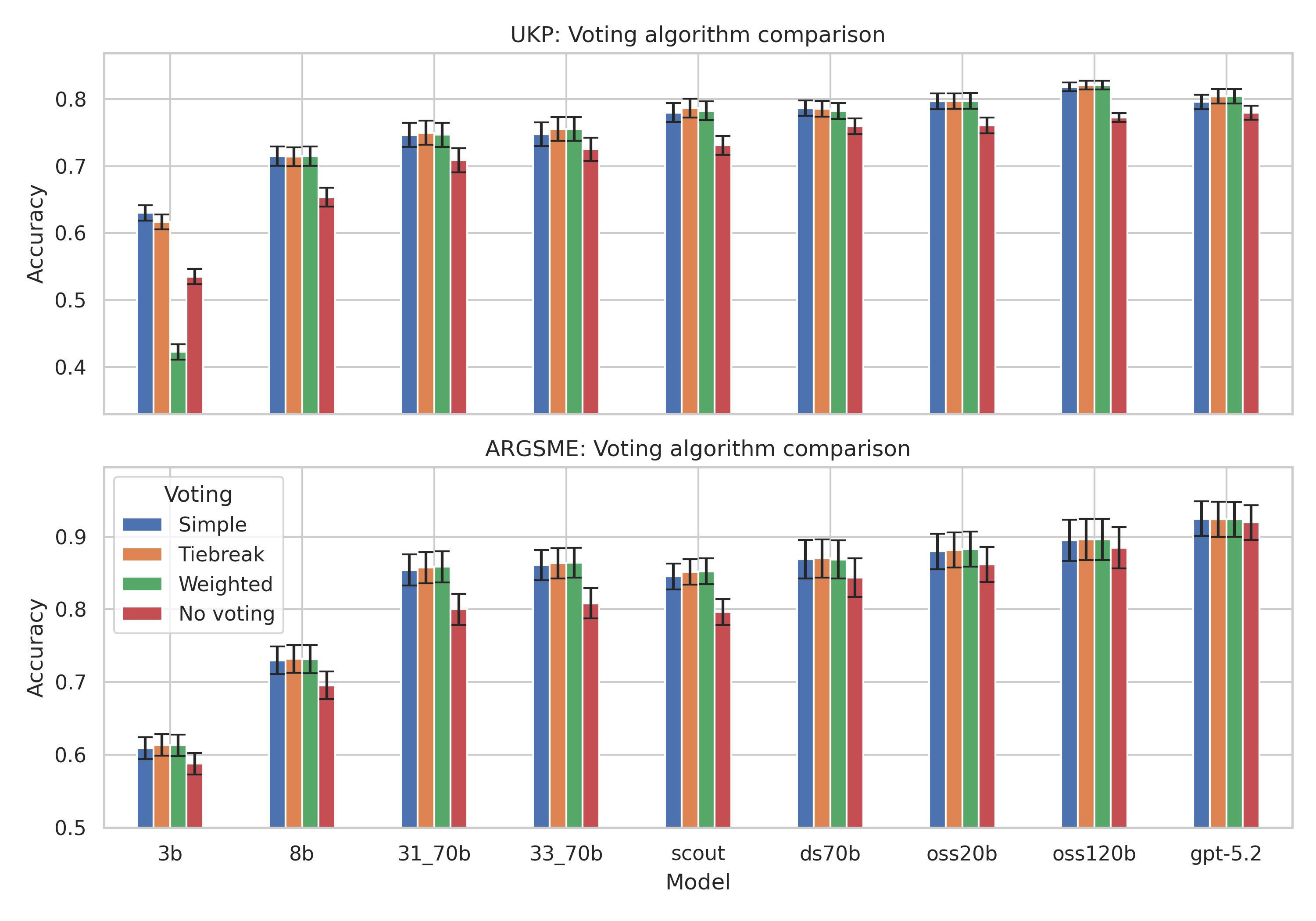}
\caption{Voting algorithm comparison (simple, tiebreak, weighted) for UKP and Args.me.}
\label{fig:voting-algorithms}
\end{figure}

\begin{table}[ht]
\centering
\caption{Average improvement in percentage points over baseline accuracy by voting algorithm}
\small
\begin{tabular}{lrr}
\toprule
\textbf{} & \textbf{UKP} & \textbf{Args.me} \\
\midrule
Simple   & +4.4\% & +2.6\% \\
Tiebreak & +4.5\% & +2.8\% \\
Weighted & +2.4\% & +2.9\% \\
\bottomrule
\end{tabular}
\label{tab:voting-algorithms}
\end{table}

We now proceed to a discussion of the results obtained from the various voting algorithms introduced in Section \ref{subsec:voting_algorithms}. A comparison against the zero-shot baseline for each model is illustrated in Figure \ref{fig:voting-algorithms}. In nearly all instances, the voting algorithms yield a statistically significant improvement over zero-shot prompting. The average percentage-point gains across different algorithms are detailed in Table \ref{tab:voting-algorithms}. Overall, the Tiebreak method achieves the most substantial improvement, followed closely by Simple voting. While Weighted voting still outperforms average zero-shot inference, it underperforms relative to the other two methods within the UKP corpus.

These results suggest that repeatedly querying a model with slight prompt variations allows to extract information than a single zero-shot trial. Consequently, reasonable aggregation methods are expected to enhance performance—a principle that aligns with recent advancements in Mixture-of-Experts (MoE) architectures \parencite{Cai2025}. Our findings also indicate that a model's self-assessment of certainty is of limited reliability. This is particularly evident in the poor performance of the weighted vote for Llama 3B on the UKP corpus; the model frequently assigned a certainty score of "0," implying a random guess despite actually outperforming the baseline. This is why more robust results are achieved using two-stage methods like "Simple" or "Tiebreak". However, even noisy indicators like self-assessed certainty appear to contain useful signal, as using these values for tie-breaking still yields an improvement over the random selection used in the Simple algorithm. Finally, the performance boost from voting is more pronounced for open-weight models than for the proprietary GPT-5.2. Notably, in the UKP corpus, this voting ensemble enables GPT oss-120B to surpass the state-of-the-art GPT-5.2.

\paragraph{Ablation study (RQ4)}
\begin{figure}[h]
\centering
\includegraphics[width=14 cm]{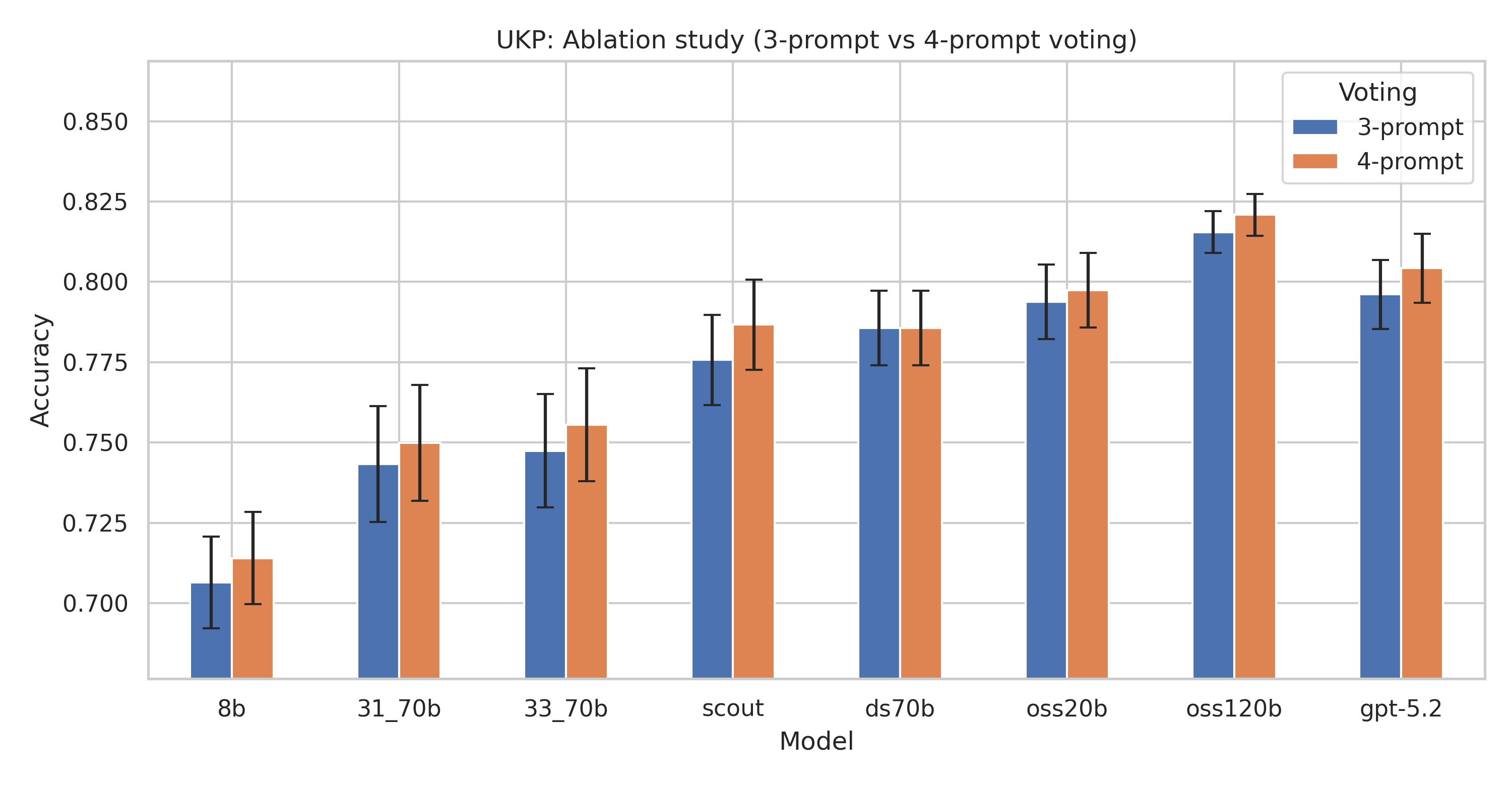}
\caption{Three-prompt vs four-prompt voting by model.}
\label{fig:ablation-ukp}
\end{figure}

\begin{table}[H]
\centering
\caption{Ablation study of the impact of presented prompts (UKP tiebreak accuracy).}
\label{tab:ablation_UKP}
\small
\begin{tabular}{lcccccc}
\toprule
\textbf{model} & \textbf{P2,3,4} & \textbf{P1,3,4} & \textbf{P1,2,4} & \textbf{P1,2,3} & \textbf{avg} & \textbf{P1,2,3,4} \\
\midrule
Llama 1b & 30.7 & 47.2 & \textbf{47.6} & 32.5 & 39.5 & 42.2 \\
Llama 3b & 59.8 & 60.9 & 60.4 & \textbf{62.7} & 60.9 & 61.7 \\
Llama 8b & 70.3 & 70.5 & 68.9 & \textbf{72.9} & 70.6 & 71.4 \\
Llama 3.1 70b & 74.1 & 73.8 & 74.4 & \textbf{75.1} & 74.3 & 75.0 \\
Llama 3.3 70b & \textbf{75.1} & 74.4 & 74.7 & 74.7 & 74.7 & 75.6 \\
Llama 4-Scout-70B & 78.1 & 77.1 & 76.3 & \textbf{78.8} & 77.6 & 78.7 \\
DS 70b & 77.9 & \textbf{79.2} & 78.4 & 78.7 & 78.6 & 78.6 \\
gpt-oss-20B & 79.3 & 79.3 & 79.3 & \textbf{79.7} & 79.4 & 79.7 \\
gpt-oss-120b & 81.5 & 81.5 & 81.2 & \textbf{81.9} & 81.5 & 82.1 \\
GPT-5.2 & 79.2 & \textbf{80.2} & 79.3 & 79.8 & 79.6 & 80.4 \\
\bottomrule
\end{tabular}
\end{table}


Given the performance variability observed across different prompts (cf. Sec. \ref{subsec:prompt_variability}), we examine whether excluding lower-performing "voters" from the ensemble can further enhance accuracy. We investigate this through an ablation study, performing voting within the four possible three-voter subsets—derived by removing one prompt at a time—using the UKP corpus. The Tiebreak algorithm identified in the previous section is employed to obtain these results.

As shown in Figure \ref{fig:ablation-ukp}, removing a voter generally leads to a degradation in average performance across models. Nevertheless, it remains to be seen whether removing a specific underperforming prompt might yield improvements. Table \ref{tab:ablation_UKP} presents the accuracy results for these configurations, with the top-performing three-prompt combination in each row highlighted in bold. The final two columns provide the mean accuracy across all three-voter groups and the original four-voter tiebreak result for reference.

For Llama 3.1 (1B, 3B, 8B, 70B), Llama 4 Scout, and the DeepSeek model, at least one three-prompt combination outperforms the four-prompt configuration. Conversely, for Llama 3.3 70B, GPT oss-20B, GPT oss-120B, and GPT-5.2, the four-prompt ensemble remains the superior choice. This suggests that maximizing the voter pool is particularly advantageous for larger models. While the P1,P2,P3 combination most frequently achieves the highest accuracy among the subsets, every combination serves as the optimal choice for at least one model. Consequently, there is no universally superior sub-selection of prompts. In summary, the most robust performance gains are achieved with a larger number of voters.

\paragraph{Performance summary and comparison with other studies (RQ2-4)}

\begin{figure}[h]
    \centering
    \includegraphics[width=14cm]{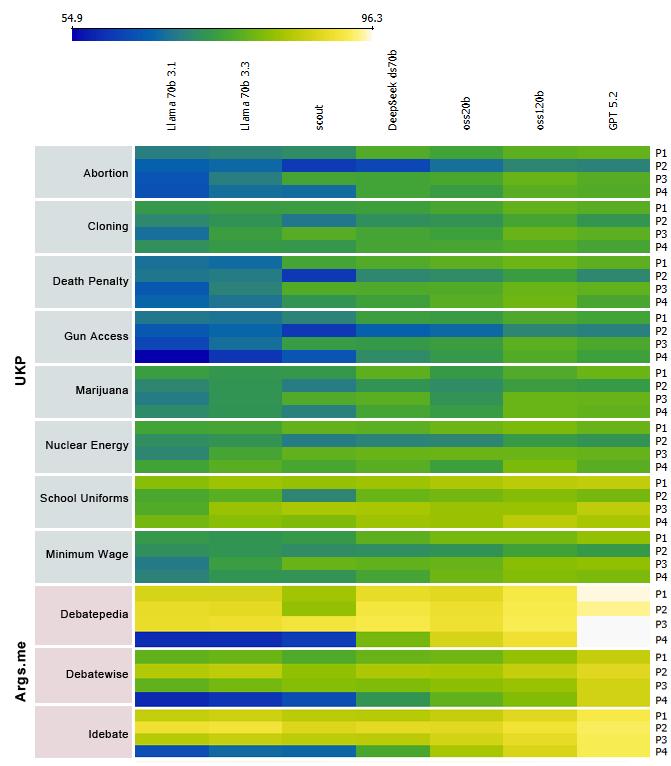}
    \caption{Heatmap visualizing the performance of the seven top-performing LLM models across debate topics in the UKP and Args.me datasets.}
    \label{heatmap1}
\end{figure}

\begin{table}[H]
\centering
\caption{Average model Accuracy in argument classification on the UKP and Args.me datasets}
\label{tab:model_results_acc}
\small
\begin{tabular}{lllllll}
\toprule
{} & \textbf{UKP (mRaR)} & \textbf{UKP (voting)} & \textbf{UKP (CoT)} & \textbf{Args.me (mRaR)} & \textbf{Args.me (voting)} & \textbf{Args.me (CoT)} \\
\midrule
Llama-3.1-1B & 37.3 $\pm$ 0.5 & 42.2 $\pm$ 0.5 & 26.6 $\pm$ 0.5 & 41.6 $\pm$ 4.1 & 41.5 $\pm$ 4.1 & 41.0 $\pm$ 4.1 \\
Llama-3.1-3B & 53.5 $\pm$ 1.1 & 61.7 $\pm$ 1.1 & 55.4 $\pm$ 1.1 & 58.8 $\pm$ 1.5 & 61.4 $\pm$ 1.5 & 55.8 $\pm$ 1.5 \\
Llama-3.1-8B & 65.4 $\pm$ 1.4 & 71.4 $\pm$ 1.4 & 65.4 $\pm$ 1.4 & 69.6 $\pm$ 1.9 & 73.2 $\pm$ 1.9 & 65.2 $\pm$ 1.9 \\
Llama-3.1-70B & 70.9 $\pm$ 1.8 & 75.0 $\pm$ 1.8 & 75.3 $\pm$ 1.8 & 80.0 $\pm$ 2.1 & 85.7 $\pm$ 2.1 & 75.8 $\pm$ 2.1 \\
Llama-3.3-70B & 72.5 $\pm$ 1.8 & 75.6 $\pm$ 1.8 & 76.8 $\pm$ 1.8 & 80.8 $\pm$ 2.1 & 86.3 $\pm$ 2.1 & 76.6 $\pm$ 2.1 \\
Llama-4-Scout-70B & 73.1 $\pm$ 1.4 & 78.7 $\pm$ 1.4 & 76.9 $\pm$ 1.4 & 79.6 $\pm$ 1.8 & 85.2 $\pm$ 1.8 & 74.2 $\pm$ 1.8 \\
ds70b & 75.9 $\pm$ 1.2 & 78.6 $\pm$ 1.2 & -- & 84.4 $\pm$ 2.6 & 87.0 $\pm$ 2.6 & -- \\
gpt-oss-20b & 76.1 $\pm$ 1.2 & 79.7 $\pm$ 1.2 & -- & 86.2 $\pm$ 2.4 & 88.2 $\pm$ 2.4 & -- \\
gpt-oss-120b & 77.2 $\pm$ 0.6 & 82.1 $\pm$ 0.6 & -- & 88.5 $\pm$ 2.8 & 89.6 $\pm$ 2.8 & -- \\
GPT-5.2 & 78.0 $\pm$ 1.1 & 80.4 $\pm$ 1.1 & -- & 91.9 $\pm$ 2.4 & 92.4 $\pm$ 2.4 & -- \\
\bottomrule
\end{tabular}
\end{table}

\begin{table}[H]
\centering
\caption{Average model F1 in argument classification on the UKP and Args.me datasets}
\label{tab:model_results_f1}
\small
\begin{tabular}{lllllll}
\toprule
{} & \textbf{UKP (mRaR)} & \textbf{UKP (voting)} & \textbf{UKP (CoT)} & \textbf{Args.me (mRaR)} & \textbf{Args.me (voting)} & \textbf{Args.me (CoT)} \\
\midrule
Llama-3.1-1B & 25.2 $\pm$ 1.4 & 35.7 $\pm$ 1.4 & 22.7 $\pm$ 1.4 & 27.8 $\pm$ 2.1 & 35.7 $\pm$ 2.1 & 30.9 $\pm$ 2.1 \\
Llama-3.1-3B & 37.5 $\pm$ 2.0 & 58.1 $\pm$ 2.0 & 38.1 $\pm$ 2.0 & 37.9 $\pm$ 0.5 & 59.1 $\pm$ 0.5 & 38.1 $\pm$ 0.5 \\
Llama-3.1-8B & 46.2 $\pm$ 1.4 & 67.2 $\pm$ 1.4 & 46.6 $\pm$ 1.4 & 45.4 $\pm$ 1.0 & 71.7 $\pm$ 1.0 & 44.0 $\pm$ 1.0 \\
Llama-3.1-70B & 52.5 $\pm$ 1.5 & 74.1 $\pm$ 1.5 & 74.2 $\pm$ 1.5 & 52.9 $\pm$ 1.3 & 84.6 $\pm$ 1.3 & 50.4 $\pm$ 1.3 \\
Llama-3.3-70B & 53.7 $\pm$ 1.4 & 74.7 $\pm$ 1.4 & 75.5 $\pm$ 1.4 & 53.5 $\pm$ 1.3 & 85.3 $\pm$ 1.3 & 76.4 $\pm$ 1.3 \\
Llama-4-Scout-70B & 52.7 $\pm$ 3.8 & 76.0 $\pm$ 3.8 & 73.0 $\pm$ 3.8 & 52.6 $\pm$ 1.4 & 83.9 $\pm$ 1.4 & 49.4 $\pm$ 1.4 \\
ds70b & 55.9 $\pm$ 1.8 & 77.0 $\pm$ 1.8 & -- & 55.8 $\pm$ 1.5 & 86.1 $\pm$ 1.5 & -- \\
gpt-oss-20b & 54.8 $\pm$ 2.8 & 76.6 $\pm$ 2.8 & -- & 57.0 $\pm$ 10.4 & 87.2 $\pm$ 10.4 & -- \\
gpt-oss-120b & 74.6 $\pm$ 1.1 & 79.8 $\pm$ 1.1 & -- & 58.3 $\pm$ 11.6 & 88.8 $\pm$ 11.6 & -- \\
GPT-5.2 & 77.3 $\pm$ 1.1 & 79.3 $\pm$ 1.1 & -- & 61.0 $\pm$ 11.9 & 91.9 $\pm$ 11.9 & -- \\
\bottomrule
\end{tabular}
\end{table}

\begin{table}[H]
\centering
\caption{Accuracy comparison across different argument mining studies}
\begin{tabular}{lrr}
\toprule
\textbf{} & \textbf{UKP (F1)} & \textbf{Args.me (Acc)} \\ 
\midrule
GPT-5.2 & 77.3 &  91.9  \\ 
gpt-oss-120b & 74.6 & 88.5 \\ 
gpt-oss-20b & 73.1 & 86.2 \\ 
ds70b & 74.6 & 84.4 \\ 
Llama 3.3 70b & 71.6 & 80.8 \\ 
Llama 3.1 8b & 61.6 & 69.6 \\ 
BERT (\cite{ref_stab18}) & 57.7 & 85.3 \\ 
LSTM (\cite{ref_stab18}) & 42.85 & - \\ 
\cite{pietron2024efficient} & 68.5 & 89.6 \\ 
\cite{akiki2020exploring} & - &   75.5 \\ 
\bottomrule
\end{tabular}

\label{tab:comparative}
\end{table}

At the end of this section we present an overview over the performance achieved by the different models using mRaR (i.e. zero-shot averaged over four prompts), Tiebreak-voting and Chain of Thought. We report Accuracy (Tab. \ref{tab:model_results_acc}) and F1 scores (Tab. \ref{tab:model_results_f1}). Detailed results such as Accuracy, Recall, Precision, F1 and sample size for all dataset, model and prompt combinations are available in Appendix \ref{appendix:results}. A visualization of model performance broken down by dataset and prompt is given in the heatmap Fig. \ref{heatmap1}.

We also present a comparison to previous studies using the UKP and Args.me benchmarks \ref{tab:comparative}. As could be expected, LLMs of the latest generation largely surpass older approaches. As a notable exception, the ensemble approach of \cite{pietron2024efficient}, which featured a combined architecture of BERT and ChatGPT-4, achieved an accuracy on par with leading models of this study for the Args.me benchmark.


\subsection{Error types in UKP dataset}

\begin{figure}[h]
\centering
\includegraphics[width=14 cm]{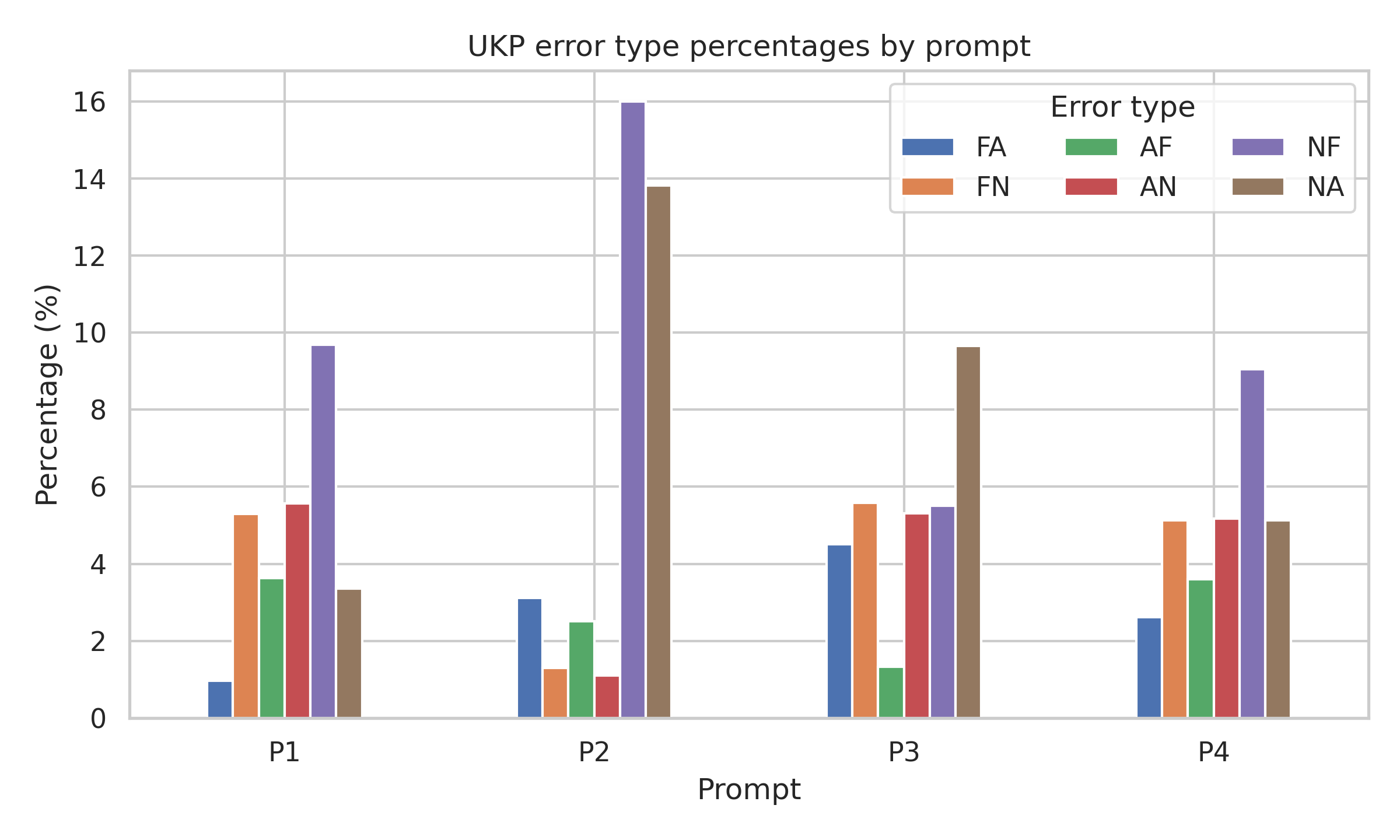}
\caption{Error types in UKP by prompt. Percentages are given with respect to all predicted arguments, including correct ones.}
\label{fig:errors-by-prompt}
\end{figure}

\begin{figure}[h]
\centering
\includegraphics[width=14 cm]{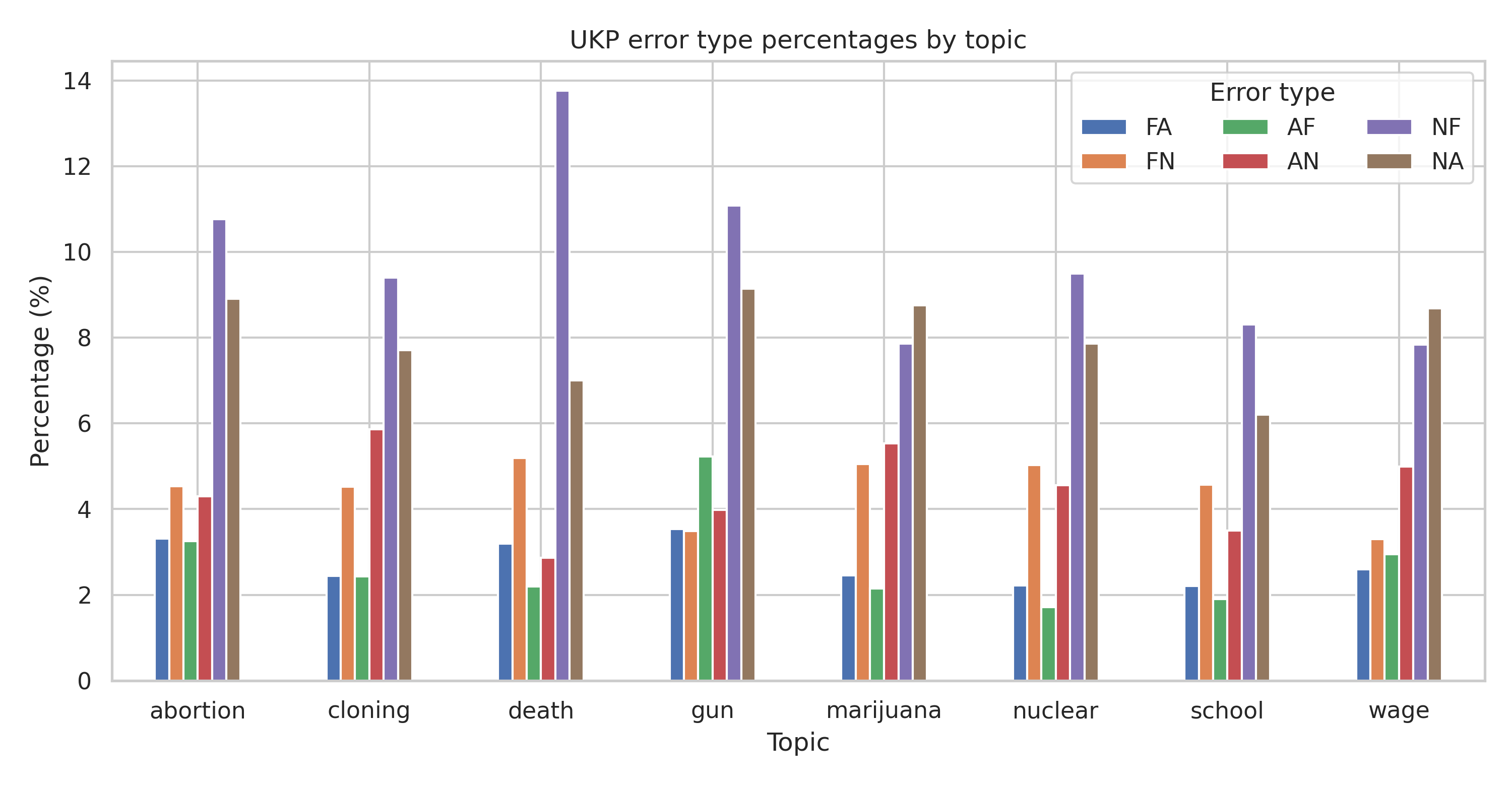}
\caption{Error types in UKP by dataset. Percentages are given with respect to all predicted arguments, including correct ones.}
\label{fig:errors-by-topic}
\end{figure}

\begin{figure}[h]
\centering
\includegraphics[width=14 cm]{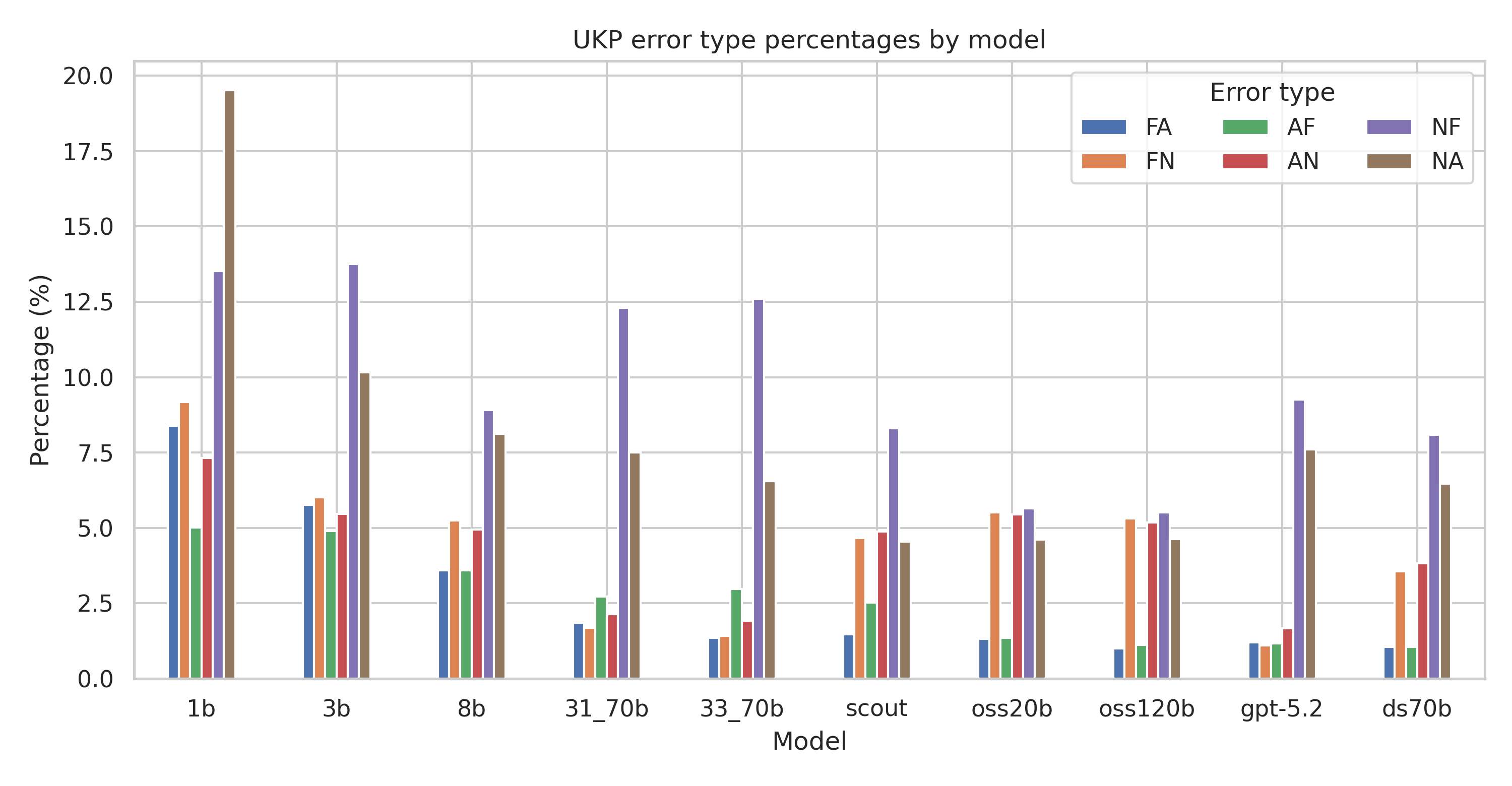}
\caption{Error types in UKP by model. Percentages are given with respect to all predicted arguments, including correct ones.}
\label{fig:errors-by-model}
\end{figure}

 Despite achieving overall good performance in argument classification, the models still make errors. Interesting patterns emerge when we analyze the types of errors that occur. Figures \ref{fig:errors-by-prompt} - \ref{fig:errors-by-model} show the proportion of the most common error types out of all errors within prompts, datasets and models, following this classification:

\begin{itemize}
\item \textbf{AF} and \textbf{AN} refer to statements labeled by human annotators as \textit{against} but incorrectly classified by the model as \textit{for} or \textit{neutral}, respectively.
\item \textbf{FA} and \textbf{FN} refer to statements labeled as \textit{for} but incorrectly classified as \textit{against} or \textit{neutral}.
\item \textbf{NA} and \textbf{NF} represent statements annotated as \textit{neutral} but misclassified by the model as \textit{against} or \textit{for}, respectively.
\end{itemize}

The most frequent type of error made by the models is classifying neutral utterances—labeled as such by annotators— making them arguments (Fig. \ref{fig:errors-by-prompt}). This suggests that the prompt’s directive to find arguments may lead the models to overinterpret content in search of argumentative structure. All prompts tend to produce NA and NF errors - that is, misclassifying neutral statements as arguments. Prompt P2, in particular, exhibits a significantly stronger tendency than the others to mislabel neutral utterances as supporting arguments (NF errors). In contrast, prompt P3 shows the most balanced distribution of error types, indicating no strong bias toward any specific classification error.

Figure \ref{fig:errors-by-topic} presents an analysis of error types by debate topic in the UKP dataset, which reveals several noteworthy patterns. When it comes to the misclassification of posts labeled as neutral by human annotators, the topics most prone to being incorrectly classified as counter-arguments (NA errors) are \textit{death penalty} and \textit{school uniforms}. In contrast, the topics most susceptible to NF errors — where neutral statements are misclassified as supporting arguments — are \textit{marijuana} and \textit{cloning}. The topic of \textit{gun access} stands out due to a notably high frequency of both FA and NA errors — where pro-access and neutral statements are misclassified as opposing arguments. This makes it the topic most prone to false negatives and suggests potential bias in the models' training data.


Among the leading models, GPT-5.2 and DS70B exhibit the strongest tilt towards NF and NA errors, meaning that they incorrectly classify neutral statements as arguments. In contrast, both gpt-oss models are significantly more balanced in terms of error types (see Fig. \ref{fig:errors-by-model}).


\section{Qualitative analysis of misclassification patterns}

In addition to quantitative evaluation, we conducted a qualitative error analysis to examine the nature of misclassifications produced by the evaluated LLMs, with particular focus on DS70B, gpt-oss-120b, and GPT-5.2, as these models achieved the highest F1 scores on the UKP dataset. The analysis aimed to identify systematic failure modes and to better understand the linguistic and argumentative conditions under which errors occur. The results revealed a range of recurring error patterns associated with discourse complexity, domain-specific characteristics, and architectural differences between the models. The findings were categorized according to the identified linguistic and structural phenomena. The results of this analysis are presented below.

\subsection{Impact of domain-specific context on error patterns (RQ5)}
\label{sec:domain-specific errors}
Error distributions varied significantly across topics. The most extreme case was the \textit{death penalty} topic, where 43.2\% of all errors were NA errors. For example: \textit{The defects in death-penalty laws, conceded by the Supreme Court in the early 1970s, have not been appreciably altered by the shift from unrestrained discretion to “guided discretion”.} This statement describes procedural developments but was misclassified as opposing the claim. Similarly, in the marijuana topic, NF errors accounted for 33.00\% of all errors. For example: \textit{For instance, cocaine has a medical purpose and can be prescribed by doctors as Erythroxylum coca, yet its true production and distribution are controlled by drug cartels and organized crime.} Although this statement provides contextual information, the oss-120b model incorrectly interpreted it as supporting the claim. These findings indicate that models tend to systematically misinterpret descriptive or contextual statements as argumentative stance.
The table below presents the dominant error type for each topic and its percentage share among all errors within that topic.

\begin{table}[ht]
\centering
\caption{Dominant error type by topic and its percentage share}
\label{tab:topic_error_distribution}

\renewcommand{\arraystretch}{1.2}
\setlength{\tabcolsep}{8pt}

\begin{tabular}{
>{\raggedright\arraybackslash}p{4cm}
>{\centering\arraybackslash}p{4cm}
>{\centering\arraybackslash}p{6cm}
}
\toprule
\textbf{Topic} & \textbf{Dominant error type} & \textbf{Percentage of errors within topic} \\
\midrule

Death penalty & NA & 43.2\% \\

Marijuana legalization & NF & 33.0\% \\

Abortion & NA & 32.3\% \\

Gun access limitation & NA & 32.3\% \\

Cloning & NA & 31.2\% \\

Minimum wage & NF & 30.3\% \\

School uniforms & NA & 29.4\% \\

Nuclear energy & NA & 28.8\% \\

\bottomrule
\end{tabular}

\end{table}

The Table \ref{tab:topic_error_distribution}  presents the dominant error type for each topic and its percentage share among all errors made by DS70B, gpt-oss-120b, and GPT-5.2 within that topic.
The analysis reveals a systematic tendency of the models to implicitly associate arguments within specific topical domains with particular argumentative positions. 
For instance, the models tend to misclassify statements as opposing, in debates on the death penalty and abortion, and as supporting in discussions on marijuana legalization and minimum wage policies. This pattern suggests that the models may implicitly associate specific topics with dominant argumentative frames, potentially reflecting biases present in their training data, and consequently assign stance based on topic-related expectations rather than the actual semantic and pragmatic content of the statement.

As certain topics appeared disproportionately associated with specific types of classification errors, we conducted an additional analysis using only the oss-120b model, which was hypothesized to exhibit topic-related bias. We examined whether, for particular topics, errors more frequently involved the misclassification of \textit{Against} arguments (AF, AN errors) than \textit{For} arguments (FA, FN errors), or vice versa. Once again, the most notable patterns emerged in datasets previously identified as particularly susceptible to model bias. For example, in the \textit{Death penalty} dataset, oss-120b reversed the polarity of \textit{For} arguments substantially more often (37.7\% of all errors) than \textit{Against} arguments (21.2\% of all errors). Moreover, when misclassifications involved statements labeled as \textit{No argument}, a large majority of them (66.7\%) were interpreted by oss-120b as opposing the death penalty. These findings suggest the presence of systematic bias in the model’s interpretation of argumentative content within this topic, leading to a disproportionate attribution of oppositional stance even when such intent was not explicitly expressed.

\subsection{Structural and discourse-level complexity in argumentation (RQ5)}

Most misclassifications across all evaluated models resulted from failures to accurately map complex syntactic structures to the correct argumentative polarity.

\begin{itemize}
    \item \textbf{Failure to interpret contrastive discourse structures}.
    The most frequent failure mode across all evaluated models involved misinterpretation of contrastive discourse structures, accounting for app. 6,876 error cases from the analysed sample. Contrastive constructions, marked by discourse operators such as \textit{but}, \textit{however}, \textit{although}, \textit{despite}, \textit{yet}, \textit{nevertheless}, or \textit{even though}, introduce a concessive clause followed by the main argumentative conclusion. In such structures, the clause following the contrastive marker typically expresses the dominant argumentative polarity, while the preceding clause provides background, qualification, or rhetorical concession. The evaluated models in some cases failed to correctly identify this hierarchical relationship. Instead, polarity was often assigned based on the first clause, while the conclusion expressed after the contrastive marker was ignored or underweighted. This resulted in systematic polarity inversion or neutralization errors, particularly in the NA (neutral misclassified as against; 2,310 cases) and NF (neutral misclassified as in favor; 2,032 cases) categories.
Example from the gun access debate (oss120b, AN): \textit{We are strong supporters of the second amendment, but we’ve got to do something to stop guns from getting into the wrong hands. The model incorrectly prioritized the concessive clause instead of the main conclusion supporting regulation.}

    \item \textbf{Misinterpretation of argument structure and multi-faceted reasoning. }Another major source of errors involved incorrect interpretation of argument structure, particularly in statements containing inference indicators such as \textit{because, therefore, since, thus, consequently}, and \textit{as a result}. These markers signal inferential relationships between premises and conclusions and are essential for identifying argumentative polarity. Correct classification requires recognizing the direction of inference and distinguishing between premises, intermediate reasoning steps, and the final evaluative conclusion. However, the evaluated models frequently failed to map these inferential structures onto the correct polarity. In app. 1,291 cases, arguments containing inference indicators were misclassified. Models often relied on surface-level lexical cues, such as references to risks, constraints, or negative consequences, without correctly interpreting their inferential role. Example form the abortion debate: (ds70b, FA): \textit{Should we ban birth control altogether then since we’d be killing a ‘human’? }The marker since introduces a premise within a rhetorical question, forming a reductio-type argument. The model misclassified the argument due to failure to interpret the inferential structure

\item \textbf{Loss of referential alignment.} Another important failure mode involved loss of referential alignment between the argument and its target claim, accounting for app. 433 cases of complete polarity reversal, primarily AF (against misclassified as in favor) and FA (in favor misclassified as against) errors. These errors were strongly associated with arguments beginning with deictic expressions such as \textit{it, this, that, these}, or \textit{those}, which function as anaphoric markers linking the statement to a previously introduced claim or context. The evaluated models frequently failed to resolve these referential dependencies, interpreting statements in isolation rather than linking pronouns to their intended targets. This resulted in polarity inversion or neutralization and reveals a limitation in discourse-level reasoning and context integration. Example form the gun access limitation debate (ds70b, FA): \textit{That’s because pulling a gun out and shooting back in the chaos of a mass shooting just makes things worse. }In this case, that refers to a previously stated claim concerning defensive gun use. The argument provides a justification opposing that claim. The model failed to correctly identify the referential target and assigned incorrect polarity.
    
\end{itemize}

\subsection{Lexical features and pragmatic inference (RQ5)}

\begin{itemize}
    \item \textbf{Literal adherence to facts.}
A total of 2,030 errors resulted from literal interpretation of empirical or statistical statements, which models classified as neutral (FN, AN), ignoring their argumentative intent. This failure mode disproportionately affected oss-120b (app. 1,222 cases), indicating a strong tendency to treat factual evidence as informational rather than argumentative. Example (oss120b, AN): \textit{A 2003 study by researchers at Arizona State University found that " students from schools without uniforms reported higher self-perception scores than students from schools with uniform policies.} Although this statement implicitly criticizes uniform policies, the model interpreted it as neutral factual information.

\item \textbf{Incorporating external discourse context}. In app. 1,146 cases, models incorporated context that was not present in the analyzed content but is often found in public debate.
Example (DS70B, FN): \textit{DNA cloning has been used in genetic engineering to create plants that offer better nutritional value}. Here, the model failed to recognize that the statement refers to a specific application of cloning and implicitly supports its permissibility in general. Instead, it incorrectly concluded that the statement does not directly address the broader ethical or social implications of cloning humans or animals and therefore classified it as not an argument.

\item \textbf{Implicit criticism, rhetorical questions, and counterfactual reasoning}.
Models frequently failed to detect implicit evaluative intent, including rhetorical questions and counterfactual constructions. This included app. 813 cases of undetected implicit criticism and app. 1,392 cases involving counterfactual or ironic reasoning. Example (oss-120b, AN): \textit{If the students do n’t like them, why should we force them to wear these garments anyway?}. This rhetorical question implicitly opposes uniform policies but was misclassified due to failure to recognize implicit argumentative intent.

\item \textbf{Attributing meaning beyond textual evidence. }In app. 685 cases, models attributed argumentative intent that was not supported by the original text, leading to polarity reversals. Example (oss-120b, AF): \textit{Some abortions happen because of societal pressures.} This statement implicitly criticizes abortion by emphasizing external coercion, yet oss-120b classified it as an argument in favor of abortion. This suggests that the model relied on generalized discourse associations rather than the actual evaluative direction expressed in the statement. 

\item \textbf{Negation and emotional language.}
Negation and emotionally charged language also disrupted classification. In app. 154 cases, sentences containing multiple negations caused polarity misclassification, particularly in oss120b and ds70b. Additionally, 617 cases involved emotionally expressive informal language that models failed to interpret correctly. Example (ds70b, AN): \textit{"The fact that some states or countries which do not use the death penalty have lower murder rates than jurisdictions which do is not evidence of the failure of deterrence"} Here, multiple negations and emotional emphasis interfered with polarity detection, therefore the model is overlooking the final negation present in the argument, concluding that this statement is evidence of the failure of deterrence.
\end{itemize}

These findings demonstrate that pragmatic inference represents a major limitation in current LLM-based argument mining. Models correctly interpret literal semantic content but sometimes  fail to infer argumentative intent, particularly when arguments rely on empirical evidence, rhetorical structure, implicit criticism, emotional emphasis, or contextual reference.

\subsection{Model-Specific Differences in Error Patterns (RQ5)}

Although all evaluated models exhibited similar general limitations related to discourse complexity and pragmatic inference, some differences emerged in their susceptibility to specific linguistic and argumentative phenomena. The observed phenomena are presented in Table \ref{tab:model_specific_weaknesses}.
Overall, these findings indicate a clear trade-off between pragmatic sensitivity and structural reasoning. The gpt-oss120b model struggles primarily with pragmatic inference and implicit intent, GPT-5.2 demonstrates strong pragmatic understanding but remains vulnerable to structural complexity, and ds70b occupies an intermediate position, with particular sensitivity to emotional language. 

\begin{table}[ht]
\centering
\caption{Model-specific characteristics in argument misclassification}
\label{tab:model_specific_weaknesses}
\begin{tabular}{p{1.5cm} p{6cm} p{8cm}}
\toprule
\textbf{Model} & \textbf{Major misclassification categories} & \textbf{Model-specific characteristics} \\
\midrule

\textbf{oss120b} &
\begin{itemize}[nosep, leftmargin=*, before=\vspace{-0.6\baselineskip}]
\item Contrastive discourse structures 
\item Literal interpretation of facts 
\item Complex argumentation 
\item Implicit criticism 
\end{itemize}
&
More vulnerable to literal interpretation and failure to recognize argumentative intent. This model more frequently misclassifies statements containing numbers, statistics, and research findings as purely neutral or informational content. It performs worse in detecting irony, rhetorical questions, and implicit criticism.
\\ \addlinespace 

\textbf{gpt-5.2} &
\begin{itemize}[nosep, leftmargin=*, before=\vspace{-0.6\baselineskip}]
\item Contrastive discourse structures 
\item Complex argumentation 
\item Misinterpretation of argument structure 
\end{itemize}
&
High resistance to literal interpretation, but occasional difficulty in processing argument structure. This model performs best in inferring argumentative intent from statistical evidence and rarely misclassifies factual arguments. However, its main limitation involves multi-faceted and contrastive sentences, where it may fail to identify the main conclusion. It may occasionally overinterpret logical structure markers.
\\ \addlinespace

\textbf{ds70b} &
\begin{itemize}[nosep, leftmargin=*, before=\vspace{-0.6\baselineskip}]
\item Contrastive discourse structures 
\item Literal interpretation of facts 
\item Complex argumentation
\item Incorporation of external context 
\item Emotional language misinterpreted as argument
\end{itemize}
&
More susceptible to emotional and informal language. ds70b demonstrates better recognition of argumentative intent in statistical statements than gpt-oss120b, but performs worse than GPT-5.2. It may misinterpret emotional expressions as indicative of argumentative stance and exhibits difficulties in correctly handling references to external discourse context.
\\
\bottomrule
\end{tabular}
\end{table}

\subsection{Errors made by LLMs and borderline cases (RQ6)}

\begin{figure}[h]
\centering
\includegraphics[width=12 cm]{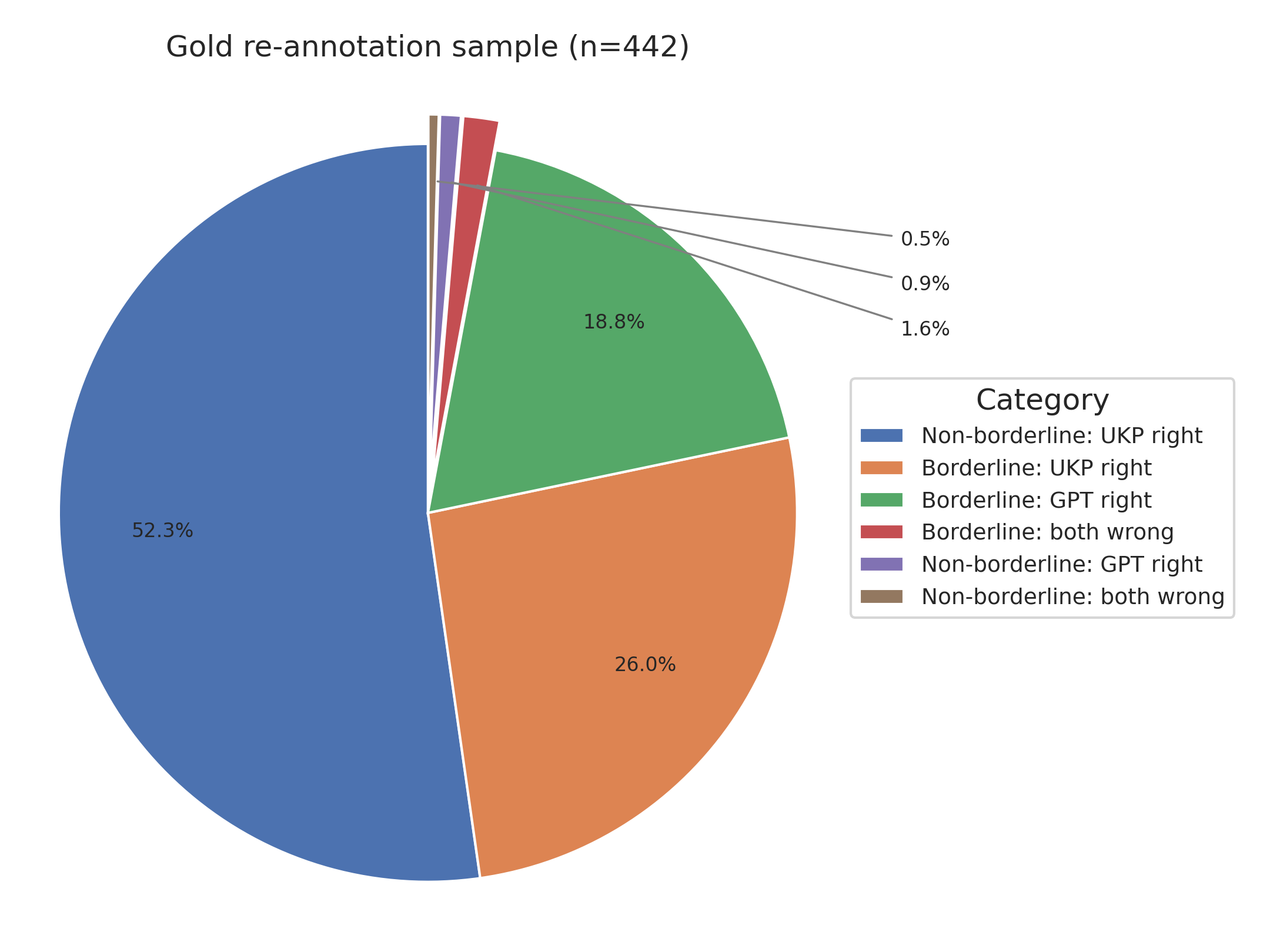}
\caption{Gold re-annotation sample breakdown (n=442).}
\label{fig:annotation-review}
\end{figure}

The accuracy of the models was assessed on the basis of discrepancies between their evaluations and those produced by the annotators involved in creating the datasets used in this research.
In other words, the values presented in the tables represent the degree of agreement between the models’ judgements and those of the annotators.

It should be noted, however, that the test datasets were constructed from real instances of argumentative exchange and, as a consequence, necessarily include borderline cases that do not lend themselves to straightforward classification.
This raises the question of the extent to which the calculated inaccuracy rates of the models result from disagreements concerning the annotation of precisely such cases.
A comprehensive answer to this question would require a meticulous analysis of the datasets employed, which lies well beyond the scope of the present study.
We therefore conducted only a limited pilot investigation of this issue.
Its results cannot be regarded as statistically significant and therefore warrant only a single, modest conclusion: namely, that the quantitative results reported in this paper may not fully reflect the models’ actual performance with respect to the assigned task.

Our preliminary investigation focused exclusively on records from the UKP \textit{abortion} dataset for which, using prompt 3, the best-performing model (GPT-5.2) produced 442 responses that diverged from those of the annotators (Fig. \ref{fig:annotation-review}).
This particular dataset–prompt combination was selected because GPT-5.2’s results for it were closest to the model’s average performance on the UKP collection (78.0\%).
An examination of all these cases revealed that a substantial proportion - 205 instances (46.4\%) - consisted of borderline expressions that could plausibly be classified differently from the way they were originally annotated.
Our annotator not only identified the problematic cases but also independently re-annotated all 442 instances.
It transpired that 90 out of the 205 borderline cases (43.9\%) were annotated differently from the original UKP labels.
Of these 90 cases, 83 (92.2\%) were classified in accordance with the model’s output.
By contrast, among the 237 cases not identified as borderline, only 6 (2.5\%) were annotated differently from the UKP dataset; in 4 of these cases, our classification again coincided with that of the model.

It should be emphasised that the vast majority of divergently classified records—355 cases (80.3\%)—were nevertheless judged to constitute genuine model inaccuracies.
However, if our limited observations were to be confirmed by statistically robust studies, they would suggest that the actual accuracy of the models may be somewhat higher than indicated by our current results.
Any further investigation would also need to consider records for which the models’ outputs were deemed “correct,” since annotation errors in such cases would in fact increase the number of genuine model inaccuracies.
One may conjecture that within cases of agreement there are significantly fewer borderline expressions and hence fewer annotation errors (the low proportion of disagreement among non-borderline cases appears to support this assumption).
At present, however, this conjecture lacks sufficient empirical support and cannot serve as a reliable basis for definitive conclusions.



\section{Conclusions and future work}

The findings of this study suggest that argument classification represents a particularly sensitive benchmark for evaluating the reasoning capabilities of large language models. Our results also highlight the importance of complementing standard quantitative evaluation with systematic qualitative error analysis. While aggregate accuracy and F1 scores suggest substantial progress in argument classification, qualitative inspection reveals persistent structural limitations shared across models, including failures in referential resolution, contrastive reasoning, and pragmatic inference. These findings suggest that future benchmarks should incorporate evaluation protocols specifically designed to test discourse-level reasoning, rather than relying solely on aggregate classification performance. In particular, the inclusion of adversarial examples, contrastive constructions, counterfactual statements and pragmatically implicit arguments may provide a more accurate assessment of true reasoning capability.

An important practical implication of this study is that a substantially smaller open-weight model, such as gpt-oss-120b, can perform only marginally worse than the flagship proprietary model GPT-5.2 in argument mining tasks. When enhanced with structured prompting and certainty-based voting strategies, the performance gap narrows further, in some configurations approaching parity. This result demonstrates that near state-of-the-art argument classification does not necessarily require the largest or proprietary architectures. From a computational perspective, this finding is particularly relevant: achieving comparable performance with a smaller open model translates into significantly lower infrastructure demands, reduced server load, and improved cost-efficiency. Consequently, high-quality argument mining systems can be developed in a more accessible, scalable, and reproducible manner, without exclusive dependence on closed commercial models.

These observations are further supported by our pilot re-annotation study conducted on 442 disagreement cases in the UKP abortion dataset. Nearly half of these instances (46.4\%) were identified as borderline expressions, and in 43.9\% of those cases the re-annotation differed from the original labels—most often aligning with the model’s prediction. This indicates that a non-negligible proportion of reported “model errors” may in fact reflect annotation ambiguity rather than genuine reasoning failure. Consequently, improving dataset quality should not only involve stricter annotation procedures but also clearer formalization of borderline categories, including explicit guidelines on the treatment of implicit arguments, rhetorical questions, descriptive-statistical statements, and context-dependent evaluative claims. Transparent publication of annotation policies and decision criteria would enhance reproducibility, allow fairer model comparison, and provide a more reliable benchmark for evaluating discourse-level reasoning in LLMs.

The limitations identified in this study have important implications for the deployment of LLM-based argument mining systems in real-world contexts, such as public policy analysis, legal reasoning, and social media monitoring. In particular, the tendency of models to infer argumentative intent based on topic associations rather than textual evidence raises concerns about potential bias amplification and misrepresentation of viewpoints. This suggests that fully automated argument classification systems should be used cautiously in high-stakes contexts and should ideally be complemented by human oversight or hybrid human–AI annotation workflows. The observed failure modes indicate that further progress in argument mining will likely require architectural and training advances specifically targeting discourse-level reasoning. In particular, improvements may be achieved through training regimes that explicitly incorporate discourse structure, argumentation theory, and pragmatic inference, rather than relying solely on next-token prediction objectives.

Overall, these results confirm that while modern LLMs represent a major advancement in automated argument mining, their limitations in pragmatic reasoning and discourse-level interpretation remain a fundamental challenge. Addressing these shortcomings will require improvements not only in model architecture but also in training data quality, annotation frameworks, and reasoning-oriented training methods.

Based on our results, future work might focus on developing more sophisticated prompt engineering-based algorithms that improve argument classification. The basic binary and three-label argument classification might be extended to more complex argument mining tasks. Another plausible next step would be to adapt the RAG technique and fine tuning of models such as gpt-oss-20b for accurate and highly efficient argument mining.



\printbibliography
\appendix
\section{Models used in the study}
\label{appendix:models}
\begin{table}[ht]
\centering
\small
\begin{tabular}{lllccr}
\toprule
\textbf{Model} & \textbf{Developer} & \textbf{Release date} & \textbf{Huggingface code} & \textbf{License}  & \textbf{Code}\\
\midrule
Llama 3.2 1B & Meta & Sep 2024 & meta-llama/Llama-3.2-1B-Instruct & open & 1b \\
Llama 3.2 3B & Meta & Sep 2024 & meta-llama/Llama-3.2-3B-Instruct & open & 3b \\
Llama 3.1 8B & Meta & Jul 2024 & meta-llama/Llama-3.1-8B-Instruct & open & 8b \\
Llama 3.1 70B & Meta & Jul 2024 & meta-llama/Llama-3.1-70B-Instruct & open & 31\_70b \\
Llama 3.3 70B & Meta & Dec 2024 & meta-llama/Llama-3.3-70B-Instruct & open & 33\_70b \\
Llama 4 Scout & Meta & Apr 2025 & meta-llama/Llama-4-Scout-17B-16E-Instruct & open & scout \\
DeepSeek R1 Dist. 70B & DeepSeek & Jan 2025 & deepseek-ai/DeepSeek-R1-Distill-Llama-70B & open & ds70b \\
gpt-oss-20B & OpenAI & Aug 2025 & openai/gpt-oss-20b & open & oss20b \\
gpt-oss-120b & OpenAI & Aug 2025 & openai/gpt-oss-120b & open & oss120b \\
GPT-5.2 & OpenAI & Dec 2025 & -- & proprietary & gpt-5.2 \\

\bottomrule
\end{tabular}
\label{tab:models}
\end{table}

The models in this study are summarized in the table below. The last column represents the code which was used in this paper to refer to the models within figures and tables.

\section{Prompts}
\label{appendix:prompts}
We used several distinct prompts to query the models, varying in complexity and expected output format. The prompts differed slightly between the UKP and Args.me datasets due to the different nature of the corpora (UKP includes a "No Argument" class, whereas Args.me is binary). In addition, for each prompt 1-4, a version with certainty estimation was also used. In each case, we used the standard system prompt (if provided) and chat template for the given model. 

\subsection*{UKP}

All prompts for the UKP dataset operate on a sentence and a thesis.

\begin{table}[H]
\centering
\small
\begin{tabular}{l p{14cm}}
\toprule
\textbf{No.} & \textbf{Text} \\
\midrule
1 & Is the sentence: "\texttt{\{sentence\}}" an argument for or against \texttt{\{thesis\}}, or is it no argument? Return one of the expressions: ``For'', ``Against'' or ``No argument'', without any additional commentary. \\ \addlinespace
2 & The thesis is: "\texttt{\{thesis\}}" Indicate if the argument "\texttt{\{sentence\}}" is for this thesis (F), against this thesis (A), or neutral (N). Please respond with only one letter: F, A, or N, without any additional commentary. \\ \addlinespace
3 & In the context of the ongoing public debate, evaluate whether the text "\texttt{\{sentence\}}" represents an argument supporting or opposing "\texttt{\{thesis\}}", or whether it does not qualify as an argument at all. Respond with one of the expressions: ``For'', ``Against'' or ``No Argument''. \\ \addlinespace
4 & Is the sentence: "\texttt{\{sentence\}}" an argument for (F) or against (A) \texttt{\{thesis\}}, or is it no argument (N)? Return a single letter: F, A, or N, without any additional commentary. \\
\bottomrule
\end{tabular}
\end{table}

Below are the {thesis} variants for the eight UKP datasets. For Prompt 2 we used the elaborate "long" formulation, for Prompts 1, 3 and 4 the "short" formulations.

\begin{table}[H]
\centering
\small
\begin{tabular}{lll}
\toprule
\textbf{Dataset} & \textbf{Short \texttt{\{thesis\}}} & \textbf{Long \texttt{\{thesis\}}} \\
\midrule
Abortion & abortion & Abortion should be fully accessible. \\
Cloning & cloning & Cloning should be allowed. \\
Death penalty & death penalty & The death penalty should be allowed. \\
Marijuana & legalisation of marijuana & Marijuana should be legal. \\
Gun laws & stricter gun laws & Gun access should be limited. \\
Minimum wage & minimum wage & The minimum wage is justified and should be increased. \\
Nuclear energy & nuclear energy & Nuclear energy should be developed. \\
School uniforms & school uniforms & School uniforms should be the standard in education. \\
\bottomrule
\end{tabular}
\end{table}

\subsection*{Args.me}

For the Args.me dataset, the "No Argument" option was removed.

\begin{table}[H]
\centering
\small
\begin{tabular}{l p{14cm}}
\toprule
\textbf{No.} & \textbf{Text} \\
\midrule
1 & Is the sentence: "\texttt{\{sentence\}}" an argument for or against "\texttt{\{thesis\}}"? Return one of the expressions: "For" or "Against", without any additional commentary. \\ \addlinespace
2 & The thesis is: "\texttt{\{thesis\}}" Indicate if the argument "\texttt{\{sentence\}}" is for this thesis (F) or against this thesis (A). Please respond with only one letter: F or A, without any additional commentary. \\ \addlinespace
3 & In the context of the ongoing public debate, evaluate whether the text "\texttt{\{sentence\}}" represents an argument supporting or opposing "\texttt{\{thesis\}}". Respond with one of the expressions: "For" or "Against". \\ \addlinespace
4 & Is the sentence: "\texttt{\{sentence\}}" an argument for (F) or against (A) \texttt{\{thesis\}}? Return a single letter: F or A, without any additional commentary. \\  \addlinespace
\bottomrule
\end{tabular}
\end{table}

\subsection*{Certainty}
\label{appendix:certainty_prompt}
For the certainty self-rating, the model was first queried with one of the above prompts (1-4). Then, the following follow-up prompt was appended to the conversation history:

\begin{quote}
"Return the certainty of your answer as a percentage. Output only a single number between 0 and 100, with no additional text."
\end{quote}

\subsection*{Chain of Thought}
\label{appendix:reasoning_prompt}
To evaluate reasoning capabilities of the Llama models, we used a Chain of Thought (CoT) prompt. This prompt explicitly instructs the model to think step-by-step.

\begin{table}[H]
\centering
\small
\begin{tabular}{l p{13cm}}
\toprule
\textbf{Corpus} & \textbf{CoT Prompt} \\
\midrule
UKP & Is the sentence: "\texttt{\{sentence\}}" an argument for or against \texttt{\{thesis\}} or is it no argument?
\newline
Solve the argument classification problem. Think through the problem step by step to solve it.
\newline
Then output one final line exactly in the format: FINAL: <For|Against|No Argument>
\newline
Do not include anything after the FINAL line. \\ \addlinespace
Args.me & Is the sentence: "\texttt{\{sentence\}}" an argument for or against \texttt{\{thesis\}}?
\newline
Solve the argument classification problem. Think through the problem step by step to solve it.
\newline
Then output one final line exactly in the format: FINAL: <For|Against>
\newline
Do not include anything after the FINAL line.\\ \addlinespace
\bottomrule
\end{tabular}
\end{table}

\section{Experimental Setup and Hyperparameters}

All local model inferences (Llama models, DeepSeek 70B, gpt-oss 20b, and gpt-oss 120b) were conducted on the Athena supercomputer at the Academic Computer Centre Cyfronet AGH. The computations were performed on nodes equipped with 8 $\times$ NVIDIA A100 GPUs (40GB VRAM each).

Local models were deployed using the vLLM library. We utilized the default vLLM sampling parameters, with the exception of temperature, which was set to $0.6$ for all models. The maximum number of new tokens generated was set to 4096. Other hyperparameters such as reasoning effort for the GPT models were also left to the default values.

Inference for GPT-5.2 was performed via cloud inference using the OpenRouter API. The total expenditure for cloud inference amounted to USD 300 and was funded by the Department of Humanities, AGH University of Science and Technology.

\section{Answer Parsing}
\label{appendix:parsing}
To extract the final classification from the model outputs, we applied regular expressions (Regex) designed to capture the expected label formats while ignoring extraneous text (e.g., "The answer is..."). The regex patterns generally looked for the keywords at the beginning or end of the string, or emphasized within the text (e.g., inside quotes or asterisks).

Below are the regex patterns used for parsing the cleaned model outputs (whitespace and punctuation removed, converted to lowercase):

\begin{table}[H]
\centering
\small
\begin{tabular}{l l p{13cm}}
\toprule
\textbf{Corpus} & \textbf{Prompt No.} & \textbf{Pattern} \\
\midrule
UKP & 1, 3 &  \verb!r'^\b(for|against|no argument)\b|\b(for|against|no argument)\b$'!\\ 
UKP & 2, 4 & \verb!r'^\b(f|a|n)\b|\b(f|a|n)\b$'!  \\ 
Args.me & 1, 3 & \verb!r'^\b(for\|against)\b|\b(for\|against)\b$'! \\  
Args.me & 2, 4 & \verb!r'^\b(f|a)\b|\b(f|a)\b$'! \\ 
\bottomrule
\end{tabular}
\end{table}

If a direct match was not found using the patterns above, a fallback search was performed to find emphasized keywords (e.g., inside quotes or asterisks):
\begin{verbatim}
r"(?:'|\"|\*\*)(\s*)(for|against|no argument)(\s*)(?:'|\"|\*\*)"
\end{verbatim}

The certainty value was obtained by searching for the pattern \verb!r'\d+'!, i.e. extracting the first number found in the output.

\newpage
\section{Detailed Results by Prompt}
\label{appendix:results}
We present the detailed results for each run. 

\begin{table}[H]
\centering
\begin{tabular}{l p{10cm}}
n & Sample size \\
Acc & Accuracy \\
Prec & Precision \\
Rec & Recall \\
F1 & F1 score (macro average) \\
Inv.\ ans. & Invalid answer: answers which were empty (mostly API issues) or could not be parsed according to the rules in App. \ref{appendix:parsing} \\
Inv.\ cert. & Invalid certainty (same as above) \\
\end{tabular}
\end{table}

\begin{longtable}{l l l l r r r r r r r}

\toprule
Corpus & Model & Dataset & Prompt & n & Acc & Prec & Rec & F1 & Inv. ans. & Inv. cert. \\
\midrule
\endfirsthead
\toprule
Corpus & Model & Dataset & Prompt & n & Acc & Prec & Rec & F1 & Inv. ans. & Inv. cert. \\
\midrule
\endhead
\bottomrule
\endfoot
Argsme & 1b & debatepedia & 1 & 21193 & 25.7 & 55.1 & 50.1 & 20.7 & 25 & 224 \\
Argsme & 1b & debatepedia & 2 & 21193 & 41.5 & 53.1 & 53.3 & 41.5 & 2 & 17259 \\
Argsme & 1b & debatepedia & 3 & 21193 & 24.5 & 52.2 & 51.5 & 31.1 & 4785 & 137 \\
Argsme & 1b & debatepedia & 4 & 21193 & 50.2 & 54.0 & 55.1 & 48.7 & 5 & 13266 \\
Argsme & 1b & debatepedia & CoT & 2000 & 43.0 & 55.7 & 57.2 & 52.3 & 410 & - \\
Argsme & 1b & debatewise & 1 & 13751 & 41.1 & 56.1 & 50.2 & 29.8 & 40 & 120 \\
Argsme & 1b & debatewise & 2 & 13751 & 47.3 & 50.2 & 50.2 & 47.0 & 1 & 10142 \\
Argsme & 1b & debatewise & 3 & 13751 & 35.6 & 50.6 & 50.3 & 39.0 & 2576 & 139 \\
Argsme & 1b & debatewise & 4 & 13751 & 50.1 & 49.8 & 49.8 & 49.5 & 8 & 5379 \\
Argsme & 1b & debatewise & CoT & 2000 & 39.2 & 52.1 & 52.2 & 51.6 & 489 & - \\
Argsme & 1b & idebate & 1 & 13048 & 50.5 & 54.7 & 50.2 & 35.0 & 81 & 102 \\
Argsme & 1b & idebate & 2 & 13048 & 51.1 & 51.0 & 50.9 & 49.4 & 0 & 9669 \\
Argsme & 1b & idebate & 3 & 13048 & 46.0 & 55.5 & 53.2 & 48.0 & 1845 & 132 \\
Argsme & 1b & idebate & 4 & 13048 & 51.2 & 51.4 & 51.4 & 51.4 & 46 & 6435 \\
Argsme & 1b & idebate & CoT & 2000 & 40.7 & 56.0 & 56.0 & 55.9 & 545 & - \\
Argsme & 3b & debatepedia & 1 & 21193 & 51.0 & 63.1 & 64.4 & 50.9 & 0 & 246 \\
Argsme & 3b & debatepedia & 2 & 21193 & 65.9 & 63.1 & 67.1 & 62.4 & 2 & 38 \\
Argsme & 3b & debatepedia & 3 & 21193 & 65.3 & 59.6 & 61.5 & 59.8 & 227 & 216 \\
Argsme & 3b & debatepedia & 4 & 21193 & 62.4 & 51.7 & 51.7 & 51.7 & 0 & 54 \\
Argsme & 3b & debatepedia & CoT & 2000 & 55.8 & 63.2 & 67.4 & 58.6 & 149 & - \\
Argsme & 3b & debatewise & 1 & 13751 & 52.2 & 58.0 & 56.5 & 51.2 & 0 & 300 \\
Argsme & 3b & debatewise & 2 & 13751 & 59.1 & 57.1 & 56.7 & 56.7 & 13 & 84 \\
Argsme & 3b & debatewise & 3 & 13751 & 56.2 & 55.6 & 55.2 & 55.1 & 383 & 234 \\
Argsme & 3b & debatewise & 4 & 13751 & 56.5 & 52.6 & 51.9 & 50.5 & 2 & 106 \\
Argsme & 3b & debatewise & CoT & 2000 & 55.2 & 59.0 & 59.3 & 58.4 & 114 & - \\
Argsme & 3b & idebate & 1 & 13048 & 61.8 & 62.6 & 61.7 & 61.1 & 0 & 227 \\
Argsme & 3b & idebate & 2 & 13048 & 61.2 & 63.0 & 61.5 & 60.1 & 7 & 46 \\
Argsme & 3b & idebate & 3 & 13048 & 56.1 & 58.5 & 57.2 & 55.2 & 205 & 89 \\
Argsme & 3b & idebate & 4 & 13048 & 52.4 & 54.0 & 52.8 & 48.6 & 0 & 72 \\
Argsme & 3b & idebate & CoT & 2000 & 56.2 & 59.1 & 59.1 & 59.1 & 98 & - \\
Argsme & 8b & debatepedia & 1 & 21193 & 74.2 & 71.7 & 78.2 & 71.6 & 0 & 17 \\
Argsme & 8b & debatepedia & 2 & 21193 & 73.1 & 66.9 & 69.8 & 67.7 & 13 & 8 \\
Argsme & 8b & debatepedia & 3 & 21193 & 69.5 & 70.6 & 77.1 & 67.9 & 31 & 17 \\
Argsme & 8b & debatepedia & 4 & 21193 & 71.5 & 63.7 & 65.0 & 64.2 & 2 & 4 \\
Argsme & 8b & debatepedia & CoT & 2000 & 68.2 & 70.7 & 77.1 & 68.9 & 72 & - \\
Argsme & 8b & debatewise & 1 & 13751 & 66.6 & 66.5 & 67.1 & 66.3 & 1 & 52 \\
Argsme & 8b & debatewise & 2 & 13751 & 69.6 & 68.6 & 67.8 & 68.0 & 19 & 15 \\
Argsme & 8b & debatewise & 3 & 13751 & 66.4 & 69.3 & 69.4 & 67.3 & 183 & 124 \\
Argsme & 8b & debatewise & 4 & 13751 & 59.5 & 56.9 & 55.7 & 55.2 & 2 & 3 \\
Argsme & 8b & debatewise & CoT & 2000 & 62.9 & 65.0 & 65.5 & 64.6 & 58 & - \\
Argsme & 8b & idebate & 1 & 13048 & 73.1 & 73.2 & 73.1 & 73.1 & 0 & 32 \\
Argsme & 8b & idebate & 2 & 13048 & 72.2 & 74.0 & 72.4 & 71.8 & 0 & 1 \\
Argsme & 8b & idebate & 3 & 13048 & 74.9 & 75.6 & 75.0 & 74.9 & 32 & 39 \\
Argsme & 8b & idebate & 4 & 13048 & 59.1 & 62.5 & 59.4 & 56.5 & 0 & 1 \\
Argsme & 8b & idebate & CoT & 2000 & 64.5 & 67.0 & 66.9 & 66.9 & 72 & - \\
Argsme & 31\_70b & debatepedia & 1 & 21193 & 88.3 & 83.9 & 89.1 & 85.8 & 3 & 0 \\
Argsme & 31\_70b & debatepedia & 2 & 21193 & 90.2 & 86.6 & 88.6 & 87.5 & 3 & 1 \\
Argsme & 31\_70b & debatepedia & 3 & 21193 & 90.5 & 86.9 & 89.0 & 87.9 & 10 & 2 \\
Argsme & 31\_70b & debatepedia & 4 & 21193 & 58.7 & 63.1 & 66.7 & 57.6 & 0 & 2 \\
Argsme & 31\_70b & debatepedia & CoT & 2000 & 77.2 & 75.0 & 82.6 & 75.1 & 0 & - \\
Argsme & 31\_70b & debatewise & 1 & 13751 & 78.9 & 78.6 & 79.5 & 78.7 & 12 & 0 \\
Argsme & 31\_70b & debatewise & 2 & 13751 & 85.2 & 84.7 & 85.4 & 84.9 & 12 & 15 \\
Argsme & 31\_70b & debatewise & 3 & 13751 & 79.1 & 79.5 & 79.3 & 79.4 & 175 & 91 \\
Argsme & 31\_70b & debatewise & 4 & 13751 & 58.5 & 60.8 & 60.7 & 58.5 & 1 & 1 \\
Argsme & 31\_70b & debatewise & CoT & 2000 & 73.7 & 73.9 & 74.7 & 73.6 & 2 & - \\
Argsme & 31\_70b & idebate & 1 & 13048 & 86.8 & 86.8 & 86.8 & 86.8 & 1 & 0 \\
Argsme & 31\_70b & idebate & 2 & 13048 & 91.1 & 91.1 & 91.1 & 91.1 & 0 & 0 \\
Argsme & 31\_70b & idebate & 3 & 13048 & 85.6 & 86.3 & 86.0 & 85.9 & 48 & 17 \\
Argsme & 31\_70b & idebate & 4 & 13048 & 63.5 & 63.8 & 63.5 & 63.3 & 0 & 0 \\
Argsme & 31\_70b & idebate & CoT & 2000 & 76.4 & 76.9 & 76.3 & 76.2 & 0 & - \\
Argsme & 33\_70b & debatepedia & 1 & 21193 & 88.2 & 83.8 & 89.4 & 85.8 & 0 & 0 \\
Argsme & 33\_70b & debatepedia & 2 & 21193 & 89.8 & 85.9 & 88.7 & 87.1 & 0 & 0 \\
Argsme & 33\_70b & debatepedia & 3 & 21193 & 90.9 & 87.4 & 89.7 & 88.4 & 1 & 0 \\
Argsme & 33\_70b & debatepedia & 4 & 21193 & 59.2 & 64.9 & 68.8 & 58.4 & 0 & 0 \\
Argsme & 33\_70b & debatepedia & CoT & 2000 & 78.2 & 75.6 & 83.2 & 76.0 & 0 & - \\
Argsme & 33\_70b & debatewise & 1 & 13751 & 79.6 & 79.8 & 80.8 & 79.5 & 0 & 0 \\
Argsme & 33\_70b & debatewise & 2 & 13751 & 86.1 & 85.6 & 86.5 & 85.8 & 1 & 0 \\
Argsme & 33\_70b & debatewise & 3 & 13751 & 80.7 & 80.1 & 80.3 & 80.2 & 21 & 0 \\
Argsme & 33\_70b & debatewise & 4 & 13751 & 60.0 & 64.2 & 63.3 & 59.8 & 0 & 0 \\
Argsme & 33\_70b & debatewise & CoT & 2000 & 73.9 & 74.6 & 75.3 & 73.8 & 0 & - \\
Argsme & 33\_70b & idebate & 1 & 13048 & 87.5 & 87.7 & 87.5 & 87.5 & 1 & 0 \\
Argsme & 33\_70b & idebate & 2 & 13048 & 91.6 & 91.7 & 91.6 & 91.6 & 0 & 0 \\
Argsme & 33\_70b & idebate & 3 & 13048 & 86.9 & 87.2 & 87.0 & 86.9 & 4 & 0 \\
Argsme & 33\_70b & idebate & 4 & 13048 & 67.2 & 68.3 & 67.1 & 66.6 & 0 & 0 \\
Argsme & 33\_70b & idebate & CoT & 2000 & 77.6 & 78.5 & 77.5 & 77.4 & 0 & - \\
Argsme & scout & debatepedia & 1 & 21193 & 83.9 & 78.8 & 83.1 & 80.4 & 3 & 0 \\
Argsme & scout & debatepedia & 2 & 21193 & 82.8 & 77.5 & 80.8 & 78.8 & 1 & 0 \\
Argsme & scout & debatepedia & 3 & 21193 & 92.2 & 89.1 & 90.9 & 89.9 & 0 & 0 \\
Argsme & scout & debatepedia & 4 & 21193 & 61.1 & 60.4 & 63.7 & 58.3 & 0 & 0 \\
Argsme & scout & debatepedia & CoT & 2000 & 73.7 & 72.8 & 80.0 & 71.7 & 1 & - \\
Argsme & scout & debatewise & 1 & 13751 & 77.3 & 77.4 & 78.3 & 77.1 & 9 & 0 \\
Argsme & scout & debatewise & 2 & 13751 & 82.6 & 82.2 & 83.0 & 82.3 & 7 & 0 \\
Argsme & scout & debatewise & 3 & 13751 & 82.6 & 82.1 & 82.2 & 82.1 & 8 & 0 \\
Argsme & scout & debatewise & 4 & 13751 & 63.1 & 65.7 & 65.5 & 63.1 & 1 & 0 \\
Argsme & scout & debatewise & CoT & 2000 & 72.0 & 73.1 & 73.7 & 72.1 & 5 & - \\
Argsme & scout & idebate & 1 & 13048 & 86.0 & 86.0 & 86.0 & 86.0 & 0 & 0 \\
Argsme & scout & idebate & 2 & 13048 & 88.7 & 88.8 & 88.7 & 88.7 & 0 & 1 \\
Argsme & scout & idebate & 3 & 13048 & 88.5 & 88.7 & 88.6 & 88.5 & 3 & 0 \\
Argsme & scout & idebate & 4 & 13048 & 66.7 & 66.9 & 66.7 & 66.6 & 0 & 0 \\
Argsme & scout & idebate & CoT & 2000 & 76.9 & 77.5 & 76.8 & 76.7 & 0 & - \\
Argsme & ds70b & debatepedia & 1 & 2000 & 90.1 & 86.0 & 90.5 & 87.8 & 0 & 1 \\
Argsme & ds70b & debatepedia & 2 & 2000 & 92.2 & 89.0 & 91.1 & 90.0 & 0 & 4 \\
Argsme & ds70b & debatepedia & 3 & 2000 & 92.6 & 89.5 & 91.9 & 90.6 & 0 & 1 \\
Argsme & ds70b & debatepedia & 4 & 2000 & 80.6 & 76.9 & 84.1 & 78.1 & 1 & 6 \\
Argsme & ds70b & debatewise & 1 & 2000 & 79.4 & 78.7 & 78.9 & 78.8 & 0 & 1 \\
Argsme & ds70b & debatewise & 2 & 2000 & 84.7 & 84.4 & 83.8 & 84.1 & 0 & 10 \\
Argsme & ds70b & debatewise & 3 & 2000 & 81.5 & 81.1 & 80.4 & 80.7 & 1 & 1 \\
Argsme & ds70b & debatewise & 4 & 2000 & 73.5 & 72.9 & 73.6 & 73.0 & 0 & 16 \\
Argsme & ds70b & idebate & 1 & 2000 & 85.5 & 85.5 & 85.5 & 85.5 & 0 & 0 \\
Argsme & ds70b & idebate & 2 & 2000 & 89.8 & 90.0 & 89.9 & 89.9 & 1 & 16 \\
Argsme & ds70b & idebate & 3 & 2000 & 85.9 & 86.1 & 85.9 & 85.9 & 0 & 0 \\
Argsme & ds70b & idebate & 4 & 2000 & 76.7 & 76.7 & 76.7 & 76.7 & 1 & 12 \\
Argsme & oss20b & debatepedia & 1 & 2000 & 89.5 & 85.2 & 90.3 & 87.1 & 0 & 164 \\
Argsme & oss20b & debatepedia & 2 & 2000 & 90.7 & 87.6 & 88.5 & 88.0 & 2 & 247 \\
Argsme & oss20b & debatepedia & 3 & 2000 & 91.0 & 87.2 & 91.2 & 88.8 & 0 & 5 \\
Argsme & oss20b & debatepedia & 4 & 2000 & 88.2 & 83.8 & 88.0 & 85.5 & 1 & 79 \\
Argsme & oss20b & debatewise & 1 & 2000 & 80.9 & 80.2 & 80.6 & 80.4 & 0 & 112 \\
Argsme & oss20b & debatewise & 2 & 2000 & 84.4 & 84.3 & 83.2 & 83.6 & 1 & 197 \\
Argsme & oss20b & debatewise & 3 & 2000 & 82.4 & 81.9 & 81.7 & 81.8 & 0 & 3 \\
Argsme & oss20b & debatewise & 4 & 2000 & 78.8 & 78.2 & 78.0 & 78.1 & 1 & 64 \\
Argsme & oss20b & idebate & 1 & 2000 & 86.7 & 86.7 & 86.7 & 86.7 & 0 & 100 \\
Argsme & oss20b & idebate & 2 & 2000 & 89.4 & 89.7 & 89.5 & 89.4 & 0 & 206 \\
Argsme & oss20b & idebate & 3 & 2000 & 87.9 & 88.1 & 88.0 & 87.9 & 0 & 2 \\
Argsme & oss20b & idebate & 4 & 2000 & 84.4 & 84.6 & 84.4 & 84.4 & 0 & 56 \\
Argsme & oss120b & debatepedia & 1 & 2000 & 92.5 & 89.3 & 92.0 & 90.5 & 0 & 0 \\
Argsme & oss120b & debatepedia & 2 & 2000 & 92.8 & 90.8 & 90.3 & 90.5 & 0 & 3 \\
Argsme & oss120b & debatepedia & 3 & 21193 & 93.2 & 90.3 & 92.5 & 91.3 & 0 & 0 \\
Argsme & oss120b & debatepedia & 4 & 21193 & 91.1 & 87.3 & 91.0 & 88.8 & 0 & 55 \\
Argsme & oss120b & debatewise & 1 & 2000 & 82.7 & 82.2 & 82.0 & 82.1 & 0 & 2 \\
Argsme & oss120b & debatewise & 2 & 2000 & 86.7 & 86.8 & 85.5 & 86.0 & 0 & 0 \\
Argsme & oss120b & debatewise & 3 & 13751 & 83.1 & 82.5 & 82.5 & 82.5 & 0 & 0 \\
Argsme & oss120b & debatewise & 4 & 13751 & 81.5 & 80.9 & 80.9 & 80.9 & 1 & 15 \\
Argsme & oss120b & idebate & 1 & 2000 & 89.3 & 89.4 & 89.4 & 89.3 & 0 & 0 \\
Argsme & oss120b & idebate & 2 & 2000 & 91.4 & 91.7 & 91.5 & 91.4 & 0 & 0 \\
Argsme & oss120b & idebate & 3 & 2000 & 90.0 & 90.1 & 90.0 & 89.9 & 0 & 0 \\
Argsme & oss120b & idebate & 4 & 13048 & 88.5 & 88.7 & 88.5 & 88.5 & 0 & 11 \\
Argsme & gpt-5.2 & debatepedia & 1 & 2000 & 96.2 & 94.3 & 95.7 & 95.0 & 0 & 2 \\
Argsme & gpt-5.2 & debatepedia & 2 & 2000 & 94.8 & 93.3 & 92.9 & 93.1 & 0 & 0 \\
Argsme & gpt-5.2 & debatepedia & 3 & 2000 & 96.4 & 94.7 & 95.8 & 95.3 & 0 & 0 \\
Argsme & gpt-5.2 & debatepedia & 4 & 1998 & 95.1 & 92.7 & 95.0 & 93.8 & 4 & 0 \\
Argsme & gpt-5.2 & debatewise & 1 & 1999 & 86.7 & 86.3 & 86.1 & 86.2 & 2 & 4 \\
Argsme & gpt-5.2 & debatewise & 2 & 1996 & 89.3 & 89.6 & 88.4 & 88.9 & 10 & 0 \\
Argsme & gpt-5.2 & debatewise & 3 & 1998 & 87.9 & 87.5 & 87.4 & 87.5 & 4 & 2 \\
Argsme & gpt-5.2 & debatewise & 4 & 1999 & 85.9 & 85.5 & 85.4 & 85.5 & 2 & 2 \\
Argsme & gpt-5.2 & idebate & 1 & 1998 & 92.6 & 92.7 & 92.7 & 92.6 & 4 & 0 \\
Argsme & gpt-5.2 & idebate & 2 & 2000 & 93.5 & 93.7 & 93.6 & 93.5 & 0 & 0 \\
Argsme & gpt-5.2 & idebate & 3 & 2000 & 93.2 & 93.3 & 93.3 & 93.2 & 0 & 2 \\
Argsme & gpt-5.2 & idebate & 4 & 2000 & 91.6 & 91.7 & 91.6 & 91.6 & 0 & 2 \\
UKP & 1b & abortion & 1 & 3929 & 41.1 & 32.6 & 34.3 & 30.1 & 7 & 85 \\
UKP & 1b & abortion & 2 & 3929 & 25.5 & 37.5 & 34.2 & 25.6 & 0 & 0 \\
UKP & 1b & abortion & 3 & 3929 & 19.4 & 44.5 & 33.9 & 12.7 & 8 & 0 \\
UKP & 1b & abortion & 4 & 3929 & 52.4 & 31.6 & 33.7 & 31.5 & 0 & 0 \\
UKP & 1b & abortion & CoT & 2000 & 25.0 & 36.6 & 34.9 & 31.3 & 510 & - \\
UKP & 1b & cloning & 1 & 3039 & 47.1 & 44.8 & 36.0 & 31.1 & 0 & 0 \\
UKP & 1b & cloning & 2 & 3039 & 31.8 & 39.3 & 37.5 & 30.3 & 0 & 0 \\
UKP & 1b & cloning & 3 & 3039 & 23.7 & 41.9 & 33.5 & 13.6 & 1 & 0 \\
UKP & 1b & cloning & 4 & 3039 & 44.9 & 31.5 & 33.4 & 28.6 & 0 & 0 \\
UKP & 1b & cloning & CoT & 2000 & 29.5 & 40.1 & 41.0 & 37.3 & 416 & - \\
UKP & 1b & death & 1 & 3651 & 53.6 & 33.3 & 34.0 & 30.6 & 0 & 0 \\
UKP & 1b & death & 2 & 3651 & 30.6 & 36.4 & 37.2 & 29.3 & 0 & 0 \\
UKP & 1b & death & 3 & 3651 & 13.4 & 31.2 & 33.0 & 8.8 & 6 & 0 \\
UKP & 1b & death & 4 & 3651 & 52.2 & 33.3 & 34.3 & 32.1 & 0 & 0 \\
UKP & 1b & death & CoT & 2000 & 25.7 & 36.9 & 37.4 & 32.6 & 491 & - \\
UKP & 1b & gun & 1 & 3341 & 48.9 & 41.0 & 37.0 & 32.3 & 0 & 0 \\
UKP & 1b & gun & 2 & 3341 & 24.0 & 40.1 & 33.9 & 23.7 & 0 & 0 \\
UKP & 1b & gun & 3 & 3341 & 25.6 & 31.0 & 34.1 & 15.7 & 0 & 0 \\
UKP & 1b & gun & 4 & 3341 & 48.6 & 35.0 & 35.8 & 31.6 & 0 & 0 \\
UKP & 1b & gun & CoT & 2000 & 24.2 & 37.6 & 37.1 & 32.7 & 532 & - \\
UKP & 1b & marijuana & 1 & 2475 & 50.1 & 39.6 & 35.7 & 30.7 & 0 & 0 \\
UKP & 1b & marijuana & 2 & 2475 & 32.6 & 42.4 & 39.1 & 32.1 & 0 & 0 \\
UKP & 1b & marijuana & 3 & 2475 & 24.5 & 41.3 & 33.5 & 14.4 & 1 & 0 \\
UKP & 1b & marijuana & 4 & 2475 & 45.9 & 29.5 & 33.0 & 27.8 & 0 & 0 \\
UKP & 1b & marijuana & CoT & 2000 & 27.5 & 38.9 & 38.7 & 36.4 & 509 & - \\
UKP & 1b & nuclear & 1 & 3576 & 56.3 & 30.8 & 35.8 & 32.2 & 0 & 0 \\
UKP & 1b & nuclear & 2 & 3576 & 28.3 & 37.1 & 36.6 & 27.7 & 0 & 0 \\
UKP & 1b & nuclear & 3 & 3576 & 18.3 & 27.1 & 33.3 & 11.7 & 1 & 0 \\
UKP & 1b & nuclear & 4 & 3576 & 54.2 & 32.1 & 33.5 & 30.0 & 0 & 0 \\
UKP & 1b & nuclear & CoT & 2000 & 26.1 & 40.0 & 42.8 & 35.4 & 532 & - \\
UKP & 1b & school & 1 & 3008 & 52.5 & 37.4 & 35.7 & 32.6 & 0 & 0 \\
UKP & 1b & school & 2 & 3008 & 28.6 & 40.4 & 36.8 & 27.3 & 0 & 0 \\
UKP & 1b & school & 3 & 3008 & 18.9 & 28.6 & 33.6 & 11.3 & 0 & 0 \\
UKP & 1b & school & 4 & 3008 & 48.8 & 29.2 & 32.1 & 28.9 & 0 & 0 \\
UKP & 1b & school & CoT & 2000 & 27.2 & 38.6 & 41.2 & 35.5 & 477 & - \\
UKP & 1b & wage & 1 & 2473 & 51.4 & 30.9 & 36.0 & 31.3 & 0 & 0 \\
UKP & 1b & wage & 2 & 2473 & 29.9 & 36.8 & 36.9 & 30.0 & 0 & 0 \\
UKP & 1b & wage & 3 & 2473 & 24.9 & 26.4 & 33.6 & 15.3 & 1 & 0 \\
UKP & 1b & wage & 4 & 2473 & 48.1 & 31.4 & 35.5 & 31.6 & 0 & 0 \\
UKP & 1b & wage & CoT & 2000 & 27.8 & 40.7 & 41.5 & 37.2 & 521 & - \\
UKP & 3b & abortion & 1 & 3929 & 57.6 & 52.5 & 52.9 & 48.2 & 0 & 0 \\
UKP & 3b & abortion & 2 & 3929 & 35.5 & 49.1 & 51.0 & 36.0 & 0 & 0 \\
UKP & 3b & abortion & 3 & 3929 & 61.6 & 47.4 & 44.0 & 43.5 & 2 & 0 \\
UKP & 3b & abortion & 4 & 3929 & 57.4 & 47.7 & 48.5 & 47.4 & 0 & 0 \\
UKP & 3b & abortion & CoT & 2000 & 54.2 & 45.2 & 45.8 & 45.5 & 51 & - \\
UKP & 3b & cloning & 1 & 3039 & 55.8 & 60.0 & 56.1 & 52.2 & 0 & 0 \\
UKP & 3b & cloning & 2 & 3039 & 44.7 & 54.8 & 55.7 & 42.8 & 0 & 0 \\
UKP & 3b & cloning & 3 & 3039 & 51.6 & 47.1 & 42.3 & 41.0 & 2 & 0 \\
UKP & 3b & cloning & 4 & 3039 & 53.3 & 51.0 & 51.3 & 50.8 & 0 & 0 \\
UKP & 3b & cloning & CoT & 2000 & 56.8 & 56.3 & 55.7 & 56.0 & 42 & - \\
UKP & 3b & death & 1 & 3651 & 54.7 & 51.4 & 48.3 & 43.6 & 0 & 0 \\
UKP & 3b & death & 2 & 3651 & 31.1 & 46.5 & 46.1 & 28.8 & 0 & 0 \\
UKP & 3b & death & 3 & 3651 & 56.1 & 45.5 & 42.4 & 39.6 & 2 & 0 \\
UKP & 3b & death & 4 & 3651 & 52.1 & 47.2 & 47.7 & 45.2 & 1 & 1 \\
UKP & 3b & death & CoT & 2000 & 54.0 & 48.3 & 49.9 & 48.8 & 60 & - \\
UKP & 3b & gun & 1 & 3341 & 53.0 & 54.4 & 53.0 & 47.0 & 0 & 0 \\
UKP & 3b & gun & 2 & 3341 & 36.9 & 51.9 & 50.5 & 36.7 & 0 & 0 \\
UKP & 3b & gun & 3 & 3341 & 61.1 & 51.6 & 49.3 & 48.2 & 2 & 0 \\
UKP & 3b & gun & 4 & 3341 & 55.3 & 51.6 & 51.7 & 50.3 & 0 & 0 \\
UKP & 3b & gun & CoT & 2000 & 52.5 & 48.1 & 49.0 & 48.3 & 79 & - \\
UKP & 3b & marijuana & 1 & 2475 & 61.3 & 59.2 & 59.6 & 57.3 & 0 & 0 \\
UKP & 3b & marijuana & 2 & 2475 & 46.2 & 56.5 & 58.4 & 44.8 & 0 & 0 \\
UKP & 3b & marijuana & 3 & 2475 & 55.5 & 54.7 & 43.3 & 41.1 & 2 & 0 \\
UKP & 3b & marijuana & 4 & 2475 & 55.9 & 51.3 & 50.9 & 50.9 & 0 & 0 \\
UKP & 3b & marijuana & CoT & 2000 & 54.6 & 52.4 & 51.4 & 51.6 & 31 & - \\
UKP & 3b & nuclear & 1 & 3576 & 64.3 & 62.5 & 61.3 & 57.7 & 0 & 0 \\
UKP & 3b & nuclear & 2 & 3576 & 37.9 & 52.4 & 54.9 & 38.1 & 0 & 0 \\
UKP & 3b & nuclear & 3 & 3576 & 62.5 & 54.3 & 45.0 & 44.3 & 0 & 0 \\
UKP & 3b & nuclear & 4 & 3576 & 59.9 & 51.7 & 52.0 & 51.2 & 0 & 0 \\
UKP & 3b & nuclear & CoT & 2000 & 55.6 & 50.2 & 51.5 & 50.7 & 41 & - \\
UKP & 3b & school & 1 & 3008 & 67.3 & 65.0 & 63.7 & 61.7 & 0 & 0 \\
UKP & 3b & school & 2 & 3008 & 40.6 & 56.2 & 57.7 & 41.0 & 0 & 0 \\
UKP & 3b & school & 3 & 3008 & 61.3 & 60.1 & 41.7 & 39.2 & 1 & 0 \\
UKP & 3b & school & 4 & 3008 & 65.2 & 57.4 & 57.8 & 57.6 & 0 & 0 \\
UKP & 3b & school & CoT & 2000 & 58.1 & 53.4 & 51.0 & 51.5 & 45 & - \\
UKP & 3b & wage & 1 & 2473 & 59.8 & 61.2 & 61.5 & 56.8 & 0 & 0 \\
UKP & 3b & wage & 2 & 2473 & 43.3 & 54.6 & 57.4 & 42.7 & 0 & 0 \\
UKP & 3b & wage & 3 & 2473 & 61.1 & 55.8 & 48.5 & 47.7 & 2 & 0 \\
UKP & 3b & wage & 4 & 2473 & 58.2 & 54.5 & 55.0 & 53.7 & 0 & 0 \\
UKP & 3b & wage & CoT & 2000 & 57.0 & 54.7 & 55.5 & 54.9 & 60 & - \\
UKP & 8b & abortion & 1 & 3929 & 64.6 & 53.6 & 52.8 & 52.9 & 0 & 53 \\
UKP & 8b & abortion & 2 & 3929 & 60.0 & 58.1 & 63.7 & 57.5 & 8 & 2 \\
UKP & 8b & abortion & 3 & 3929 & 68.1 & 57.3 & 50.2 & 51.6 & 3 & 0 \\
UKP & 8b & abortion & 4 & 3929 & 60.3 & 50.0 & 51.5 & 50.0 & 0 & 0 \\
UKP & 8b & abortion & CoT & 2000 & 63.2 & 55.2 & 53.9 & 54.2 & 37 & - \\
UKP & 8b & cloning & 1 & 3039 & 68.8 & 68.2 & 68.2 & 67.1 & 0 & 0 \\
UKP & 8b & cloning & 2 & 3039 & 62.7 & 67.0 & 69.6 & 63.5 & 5 & 0 \\
UKP & 8b & cloning & 3 & 3039 & 66.9 & 66.6 & 60.9 & 62.3 & 4 & 0 \\
UKP & 8b & cloning & 4 & 3039 & 64.6 & 63.1 & 65.6 & 63.5 & 0 & 0 \\
UKP & 8b & cloning & CoT & 2000 & 63.0 & 64.9 & 61.8 & 62.9 & 54 & - \\
UKP & 8b & death & 1 & 3651 & 67.0 & 60.7 & 60.3 & 59.3 & 0 & 0 \\
UKP & 8b & death & 2 & 3651 & 51.3 & 58.6 & 65.1 & 50.8 & 2 & 0 \\
UKP & 8b & death & 3 & 3651 & 68.6 & 60.1 & 60.0 & 59.2 & 2 & 0 \\
UKP & 8b & death & 4 & 3651 & 58.7 & 56.7 & 59.4 & 54.4 & 0 & 0 \\
UKP & 8b & death & CoT & 2000 & 67.5 & 64.7 & 62.8 & 63.4 & 39 & - \\
UKP & 8b & gun & 1 & 3341 & 63.4 & 60.5 & 61.1 & 59.1 & 0 & 0 \\
UKP & 8b & gun & 2 & 3341 & 55.0 & 56.9 & 61.5 & 54.6 & 4 & 0 \\
UKP & 8b & gun & 3 & 3341 & 66.7 & 58.7 & 54.2 & 54.9 & 1 & 0 \\
UKP & 8b & gun & 4 & 3341 & 52.4 & 50.2 & 48.7 & 47.0 & 0 & 0 \\
UKP & 8b & gun & CoT & 2000 & 61.3 & 57.5 & 57.7 & 57.6 & 51 & - \\
UKP & 8b & marijuana & 1 & 2475 & 68.9 & 67.6 & 66.0 & 65.8 & 0 & 0 \\
UKP & 8b & marijuana & 2 & 2475 & 64.2 & 67.1 & 70.6 & 64.9 & 3 & 0 \\
UKP & 8b & marijuana & 3 & 2475 & 67.9 & 68.6 & 61.0 & 62.9 & 1 & 0 \\
UKP & 8b & marijuana & 4 & 2475 & 63.9 & 61.2 & 62.7 & 61.1 & 0 & 0 \\
UKP & 8b & marijuana & CoT & 2000 & 63.2 & 63.5 & 63.5 & 63.5 & 59 & - \\
UKP & 8b & nuclear & 1 & 3576 & 75.7 & 72.1 & 67.5 & 69.0 & 0 & 0 \\
UKP & 8b & nuclear & 2 & 3576 & 63.7 & 63.0 & 71.0 & 63.0 & 3 & 0 \\
UKP & 8b & nuclear & 3 & 3576 & 70.7 & 67.2 & 58.7 & 60.8 & 5 & 0 \\
UKP & 8b & nuclear & 4 & 3576 & 71.3 & 64.7 & 65.8 & 65.1 & 0 & 0 \\
UKP & 8b & nuclear & CoT & 2000 & 67.4 & 64.7 & 63.0 & 63.8 & 27 & - \\
UKP & 8b & school & 1 & 3008 & 75.3 & 71.9 & 72.8 & 71.4 & 0 & 0 \\
UKP & 8b & school & 2 & 3008 & 62.6 & 63.4 & 70.4 & 62.3 & 3 & 0 \\
UKP & 8b & school & 3 & 3008 & 74.1 & 73.0 & 63.8 & 65.3 & 1 & 0 \\
UKP & 8b & school & 4 & 3008 & 70.6 & 65.8 & 69.1 & 66.7 & 0 & 0 \\
UKP & 8b & school & CoT & 2000 & 72.5 & 70.3 & 69.8 & 70.0 & 35 & - \\
UKP & 8b & wage & 1 & 2473 & 72.0 & 69.0 & 70.9 & 69.1 & 0 & 0 \\
UKP & 8b & wage & 2 & 2473 & 63.2 & 65.7 & 70.8 & 63.9 & 4 & 1 \\
UKP & 8b & wage & 3 & 2473 & 72.3 & 69.7 & 65.1 & 66.7 & 6 & 0 \\
UKP & 8b & wage & 4 & 2473 & 63.0 & 60.9 & 63.9 & 60.8 & 0 & 0 \\
UKP & 8b & wage & CoT & 2000 & 65.4 & 64.0 & 63.9 & 63.8 & 46 & - \\
UKP & 31\_70b & abortion & 1 & 3929 & 69.8 & 66.9 & 71.4 & 66.8 & 0 & 0 \\
UKP & 31\_70b & abortion & 2 & 3929 & 65.8 & 64.9 & 74.5 & 64.8 & 1 & 1 \\
UKP & 31\_70b & abortion & 3 & 3929 & 67.0 & 63.7 & 70.5 & 64.6 & 55 & 0 \\
UKP & 31\_70b & abortion & 4 & 3929 & 63.9 & 58.0 & 61.2 & 57.7 & 0 & 0 \\
UKP & 31\_70b & abortion & CoT & 2000 & 73.8 & 68.7 & 73.8 & 70.6 & 0 & - \\
UKP & 31\_70b & cloning & 1 & 3039 & 74.1 & 75.2 & 76.3 & 74.2 & 0 & 0 \\
UKP & 31\_70b & cloning & 2 & 3039 & 71.5 & 73.5 & 77.9 & 71.9 & 1 & 0 \\
UKP & 31\_70b & cloning & 3 & 3039 & 71.7 & 73.1 & 75.9 & 73.5 & 77 & 0 \\
UKP & 31\_70b & cloning & 4 & 3039 & 72.8 & 73.9 & 74.5 & 72.5 & 0 & 0 \\
UKP & 31\_70b & cloning & CoT & 2000 & 76.5 & 75.7 & 77.3 & 76.2 & 0 & - \\
UKP & 31\_70b & death & 1 & 3651 & 67.6 & 70.6 & 68.9 & 66.2 & 0 & 0 \\
UKP & 31\_70b & death & 2 & 3651 & 68.7 & 67.8 & 77.5 & 67.7 & 0 & 0 \\
UKP & 31\_70b & death & 3 & 3651 & 67.4 & 66.5 & 71.3 & 66.3 & 53 & 0 \\
UKP & 31\_70b & death & 4 & 3651 & 66.3 & 67.2 & 67.7 & 63.9 & 0 & 0 \\
UKP & 31\_70b & death & CoT & 2000 & 72.5 & 71.7 & 74.6 & 71.6 & 0 & - \\
UKP & 31\_70b & gun & 1 & 3341 & 69.0 & 67.7 & 73.1 & 67.7 & 0 & 0 \\
UKP & 31\_70b & gun & 2 & 3341 & 64.7 & 66.7 & 73.0 & 64.6 & 1 & 0 \\
UKP & 31\_70b & gun & 3 & 3341 & 65.9 & 65.6 & 69.9 & 65.1 & 53 & 0 \\
UKP & 31\_70b & gun & 4 & 3341 & 54.9 & 53.1 & 52.3 & 50.4 & 1 & 0 \\
UKP & 31\_70b & gun & CoT & 2000 & 70.9 & 68.2 & 73.6 & 69.4 & 0 & - \\
UKP & 31\_70b & marijuana & 1 & 2475 & 74.6 & 73.5 & 77.2 & 74.5 & 0 & 0 \\
UKP & 31\_70b & marijuana & 2 & 2475 & 71.3 & 72.7 & 78.0 & 71.8 & 1 & 0 \\
UKP & 31\_70b & marijuana & 3 & 2475 & 73.2 & 74.5 & 78.9 & 75.2 & 63 & 0 \\
UKP & 31\_70b & marijuana & 4 & 2475 & 71.8 & 71.0 & 74.7 & 71.4 & 0 & 0 \\
UKP & 31\_70b & marijuana & CoT & 2000 & 73.8 & 72.8 & 76.7 & 74.0 & 0 & - \\
UKP & 31\_70b & nuclear & 1 & 3576 & 75.6 & 73.7 & 77.6 & 74.1 & 0 & 0 \\
UKP & 31\_70b & nuclear & 2 & 3576 & 72.5 & 70.7 & 80.0 & 71.8 & 0 & 0 \\
UKP & 31\_70b & nuclear & 3 & 3576 & 74.9 & 73.2 & 79.1 & 74.8 & 74 & 1 \\
UKP & 31\_70b & nuclear & 4 & 3576 & 75.8 & 73.1 & 77.6 & 73.9 & 0 & 0 \\
UKP & 31\_70b & nuclear & CoT & 2000 & 76.2 & 72.7 & 76.9 & 74.3 & 0 & - \\
UKP & 31\_70b & school & 1 & 3008 & 81.7 & 79.0 & 82.7 & 80.2 & 0 & 0 \\
UKP & 31\_70b & school & 2 & 3008 & 76.8 & 74.7 & 82.8 & 76.3 & 1 & 0 \\
UKP & 31\_70b & school & 3 & 3008 & 80.0 & 77.4 & 81.5 & 79.0 & 53 & 0 \\
UKP & 31\_70b & school & 4 & 3008 & 80.4 & 76.5 & 81.0 & 78.1 & 0 & 0 \\
UKP & 31\_70b & school & CoT & 2000 & 83.2 & 80.2 & 83.6 & 81.7 & 0 & - \\
UKP & 31\_70b & wage & 1 & 2473 & 74.1 & 72.5 & 76.4 & 72.8 & 0 & 0 \\
UKP & 31\_70b & wage & 2 & 2473 & 72.6 & 72.4 & 79.0 & 72.7 & 1 & 0 \\
UKP & 31\_70b & wage & 3 & 2473 & 72.5 & 72.2 & 77.2 & 73.2 & 54 & 0 \\
UKP & 31\_70b & wage & 4 & 2473 & 70.5 & 68.7 & 72.6 & 68.9 & 0 & 0 \\
UKP & 31\_70b & wage & CoT & 2000 & 75.5 & 73.0 & 76.4 & 74.3 & 0 & - \\
UKP & 33\_70b & abortion & 1 & 3929 & 71.0 & 70.4 & 71.6 & 67.8 & 0 & 0 \\
UKP & 33\_70b & abortion & 2 & 3929 & 67.0 & 65.8 & 75.3 & 65.8 & 0 & 0 \\
UKP & 33\_70b & abortion & 3 & 3929 & 70.0 & 66.0 & 72.9 & 67.0 & 6 & 0 \\
UKP & 33\_70b & abortion & 4 & 3929 & 67.7 & 60.7 & 62.5 & 60.0 & 0 & 0 \\
UKP & 33\_70b & abortion & CoT & 2000 & 73.9 & 69.0 & 72.0 & 69.9 & 0 & - \\
UKP & 33\_70b & cloning & 1 & 3039 & 74.7 & 76.4 & 77.1 & 74.9 & 0 & 0 \\
UKP & 33\_70b & cloning & 2 & 3039 & 73.0 & 74.6 & 79.1 & 73.3 & 0 & 0 \\
UKP & 33\_70b & cloning & 3 & 3039 & 75.0 & 75.4 & 79.2 & 75.6 & 13 & 0 \\
UKP & 33\_70b & cloning & 4 & 3039 & 74.1 & 75.9 & 75.2 & 73.7 & 0 & 0 \\
UKP & 33\_70b & cloning & CoT & 2000 & 78.1 & 77.5 & 78.5 & 77.8 & 0 & - \\
UKP & 33\_70b & death & 1 & 3651 & 67.2 & 73.7 & 70.0 & 67.3 & 0 & 0 \\
UKP & 33\_70b & death & 2 & 3651 & 69.6 & 69.3 & 78.9 & 68.8 & 0 & 0 \\
UKP & 33\_70b & death & 3 & 3651 & 70.6 & 72.2 & 75.6 & 70.7 & 6 & 0 \\
UKP & 33\_70b & death & 4 & 3651 & 68.7 & 73.4 & 68.4 & 66.8 & 0 & 0 \\
UKP & 33\_70b & death & CoT & 2000 & 75.0 & 74.7 & 76.5 & 74.1 & 0 & - \\
UKP & 33\_70b & gun & 1 & 3341 & 68.4 & 69.0 & 73.5 & 67.5 & 0 & 0 \\
UKP & 33\_70b & gun & 2 & 3341 & 66.4 & 68.0 & 74.5 & 66.2 & 0 & 0 \\
UKP & 33\_70b & gun & 3 & 3341 & 67.6 & 67.5 & 72.4 & 66.8 & 24 & 0 \\
UKP & 33\_70b & gun & 4 & 3341 & 60.0 & 56.6 & 56.5 & 54.5 & 0 & 0 \\
UKP & 33\_70b & gun & CoT & 2000 & 71.8 & 68.5 & 73.2 & 69.7 & 0 & - \\
UKP & 33\_70b & marijuana & 1 & 2475 & 73.6 & 73.2 & 77.3 & 73.6 & 0 & 0 \\
UKP & 33\_70b & marijuana & 2 & 2475 & 73.3 & 74.2 & 79.4 & 73.7 & 0 & 0 \\
UKP & 33\_70b & marijuana & 3 & 2475 & 73.3 & 73.9 & 78.8 & 74.0 & 17 & 0 \\
UKP & 33\_70b & marijuana & 4 & 2475 & 73.0 & 72.5 & 74.9 & 72.4 & 0 & 0 \\
UKP & 33\_70b & marijuana & CoT & 2000 & 75.2 & 74.1 & 77.8 & 75.2 & 0 & - \\
UKP & 33\_70b & nuclear & 1 & 3576 & 76.2 & 75.2 & 79.0 & 75.2 & 0 & 0 \\
UKP & 33\_70b & nuclear & 2 & 3576 & 73.2 & 71.4 & 80.6 & 72.6 & 0 & 0 \\
UKP & 33\_70b & nuclear & 3 & 3576 & 76.2 & 74.6 & 80.9 & 75.9 & 32 & 0 \\
UKP & 33\_70b & nuclear & 4 & 3576 & 78.0 & 76.1 & 76.9 & 75.4 & 0 & 0 \\
UKP & 33\_70b & nuclear & CoT & 2000 & 77.6 & 74.3 & 78.1 & 75.8 & 0 & - \\
UKP & 33\_70b & school & 1 & 3008 & 83.5 & 81.4 & 84.9 & 82.3 & 0 & 0 \\
UKP & 33\_70b & school & 2 & 3008 & 78.3 & 76.2 & 84.5 & 77.8 & 0 & 0 \\
UKP & 33\_70b & school & 3 & 3008 & 83.1 & 79.5 & 85.2 & 81.6 & 2 & 0 \\
UKP & 33\_70b & school & 4 & 3008 & 82.0 & 79.4 & 81.7 & 79.8 & 0 & 0 \\
UKP & 33\_70b & school & CoT & 2000 & 84.5 & 81.6 & 84.7 & 83.0 & 0 & - \\
UKP & 33\_70b & wage & 1 & 2473 & 73.4 & 73.2 & 76.6 & 72.4 & 0 & 0 \\
UKP & 33\_70b & wage & 2 & 2473 & 73.3 & 72.9 & 79.6 & 73.4 & 0 & 0 \\
UKP & 33\_70b & wage & 3 & 2473 & 74.9 & 74.3 & 79.9 & 74.9 & 16 & 0 \\
UKP & 33\_70b & wage & 4 & 2473 & 73.4 & 72.0 & 75.4 & 71.7 & 0 & 0 \\
UKP & 33\_70b & wage & CoT & 2000 & 78.5 & 76.1 & 79.4 & 77.4 & 0 & - \\
UKP & scout & abortion & 1 & 3929 & 72.0 & 65.4 & 57.2 & 57.9 & 0 & 0 \\
UKP & scout & abortion & 2 & 3929 & 60.5 & 62.3 & 70.4 & 59.9 & 1 & 0 \\
UKP & scout & abortion & 3 & 3929 & 76.7 & 71.2 & 74.8 & 72.7 & 0 & 0 \\
UKP & scout & abortion & 4 & 3929 & 67.4 & 53.8 & 51.3 & 52.1 & 0 & 0 \\
UKP & scout & abortion & CoT & 2000 & 74.1 & 71.4 & 62.5 & 65.2 & 0 & - \\
UKP & scout & cloning & 1 & 3039 & 74.9 & 78.8 & 69.7 & 71.6 & 0 & 0 \\
UKP & scout & cloning & 2 & 3039 & 68.9 & 72.6 & 76.0 & 69.5 & 0 & 0 \\
UKP & scout & cloning & 3 & 3039 & 78.0 & 77.4 & 79.3 & 77.8 & 0 & 0 \\
UKP & scout & cloning & 4 & 3039 & 73.8 & 75.5 & 69.8 & 71.4 & 0 & 0 \\
UKP & scout & cloning & CoT & 2000 & 75.3 & 78.8 & 70.9 & 73.0 & 0 & - \\
UKP & scout & death & 1 & 3651 & 76.1 & 76.9 & 68.2 & 70.2 & 0 & 0 \\
UKP & scout & death & 2 & 3651 & 60.4 & 65.2 & 70.5 & 60.3 & 0 & 0 \\
UKP & scout & death & 3 & 3651 & 77.7 & 76.3 & 76.4 & 75.6 & 0 & 0 \\
UKP & scout & death & 4 & 3651 & 73.3 & 67.3 & 67.1 & 67.1 & 0 & 0 \\
UKP & scout & death & CoT & 2000 & 80.4 & 81.6 & 73.8 & 76.6 & 0 & - \\
UKP & scout & gun & 1 & 3341 & 70.8 & 65.3 & 62.9 & 63.5 & 0 & 0 \\
UKP & scout & gun & 2 & 3341 & 60.5 & 63.6 & 68.8 & 60.4 & 0 & 0 \\
UKP & scout & gun & 3 & 3341 & 74.9 & 71.4 & 75.5 & 72.7 & 0 & 0 \\
UKP & scout & gun & 4 & 3341 & 64.0 & 56.4 & 56.6 & 56.2 & 0 & 0 \\
UKP & scout & gun & CoT & 2000 & 72.2 & 69.0 & 65.4 & 66.4 & 0 & - \\
UKP & scout & marijuana & 1 & 2475 & 73.9 & 76.2 & 68.3 & 69.7 & 0 & 0 \\
UKP & scout & marijuana & 2 & 2475 & 69.7 & 71.2 & 75.8 & 70.1 & 0 & 0 \\
UKP & scout & marijuana & 3 & 2475 & 77.8 & 76.4 & 79.2 & 77.5 & 0 & 0 \\
UKP & scout & marijuana & 4 & 2475 & 70.3 & 69.2 & 64.7 & 65.6 & 0 & 0 \\
UKP & scout & marijuana & CoT & 2000 & 74.6 & 76.6 & 70.2 & 72.5 & 0 & - \\
UKP & scout & nuclear & 1 & 3576 & 79.0 & 79.6 & 70.4 & 73.3 & 0 & 0 \\
UKP & scout & nuclear & 2 & 3576 & 69.7 & 68.5 & 76.9 & 69.2 & 0 & 0 \\
UKP & scout & nuclear & 3 & 3576 & 79.3 & 75.7 & 79.3 & 77.2 & 0 & 0 \\
UKP & scout & nuclear & 4 & 3576 & 76.5 & 73.3 & 69.4 & 71.0 & 0 & 0 \\
UKP & scout & nuclear & CoT & 2000 & 77.9 & 77.3 & 70.9 & 73.4 & 0 & - \\
UKP & scout & school & 1 & 3008 & 82.9 & 83.2 & 77.0 & 79.2 & 0 & 0 \\
UKP & scout & school & 2 & 3008 & 71.2 & 71.7 & 79.9 & 71.5 & 0 & 0 \\
UKP & scout & school & 3 & 3008 & 85.2 & 82.8 & 85.3 & 83.8 & 0 & 0 \\
UKP & scout & school & 4 & 3008 & 81.2 & 78.1 & 77.3 & 77.7 & 0 & 0 \\
UKP & scout & school & CoT & 2000 & 83.8 & 84.0 & 79.4 & 81.4 & 0 & - \\
UKP & scout & wage & 1 & 2473 & 74.2 & 78.0 & 64.4 & 67.8 & 0 & 0 \\
UKP & scout & wage & 2 & 2473 & 72.0 & 71.6 & 77.3 & 71.7 & 0 & 0 \\
UKP & scout & wage & 3 & 2473 & 80.1 & 78.2 & 78.8 & 78.5 & 1 & 0 \\
UKP & scout & wage & 4 & 2473 & 73.2 & 70.2 & 65.4 & 67.1 & 0 & 0 \\
UKP & scout & wage & CoT & 2000 & 76.8 & 76.9 & 71.1 & 73.2 & 0 & - \\
UKP & ds70b & abortion & 1 & 2000 & 77.5 & 75.0 & 69.7 & 71.8 & 0 & 0 \\
UKP & ds70b & abortion & 2 & 2000 & 62.1 & 62.2 & 72.2 & 61.7 & 0 & 1 \\
UKP & ds70b & abortion & 3 & 2000 & 76.0 & 71.0 & 75.4 & 72.8 & 0 & 0 \\
UKP & ds70b & abortion & 4 & 2000 & 76.0 & 71.2 & 71.1 & 71.2 & 0 & 3 \\
UKP & ds70b & cloning & 1 & 2000 & 75.3 & 76.3 & 72.5 & 73.9 & 0 & 0 \\
UKP & ds70b & cloning & 2 & 2000 & 72.5 & 72.9 & 77.7 & 73.0 & 0 & 1 \\
UKP & ds70b & cloning & 3 & 2000 & 76.2 & 75.5 & 76.2 & 75.8 & 0 & 0 \\
UKP & ds70b & cloning & 4 & 2000 & 76.3 & 76.0 & 75.5 & 75.6 & 0 & 0 \\
UKP & ds70b & death & 1 & 2000 & 77.5 & 75.7 & 74.8 & 74.8 & 0 & 0 \\
UKP & ds70b & death & 2 & 2000 & 71.5 & 68.9 & 77.1 & 70.1 & 0 & 4 \\
UKP & ds70b & death & 3 & 2000 & 77.3 & 74.6 & 76.6 & 75.0 & 0 & 0 \\
UKP & ds70b & death & 4 & 2000 & 75.4 & 74.5 & 75.4 & 74.1 & 0 & 7 \\
UKP & ds70b & gun & 1 & 2000 & 75.1 & 71.8 & 74.6 & 72.9 & 0 & 0 \\
UKP & ds70b & gun & 2 & 2000 & 65.7 & 66.5 & 73.1 & 65.4 & 0 & 2 \\
UKP & ds70b & gun & 3 & 2000 & 74.1 & 71.0 & 76.1 & 72.5 & 0 & 0 \\
UKP & ds70b & gun & 4 & 2000 & 71.8 & 68.6 & 72.3 & 69.4 & 0 & 2 \\
UKP & ds70b & marijuana & 1 & 2000 & 78.5 & 78.0 & 77.8 & 77.9 & 0 & 0 \\
UKP & ds70b & marijuana & 2 & 2000 & 73.2 & 73.4 & 78.4 & 73.7 & 0 & 0 \\
UKP & ds70b & marijuana & 3 & 2000 & 78.2 & 77.0 & 79.6 & 78.0 & 0 & 0 \\
UKP & ds70b & marijuana & 4 & 2000 & 76.1 & 74.9 & 76.1 & 75.5 & 0 & 2 \\
UKP & ds70b & nuclear & 1 & 2000 & 78.3 & 76.2 & 73.8 & 74.9 & 0 & 0 \\
UKP & ds70b & nuclear & 2 & 2000 & 70.8 & 69.1 & 78.6 & 70.3 & 0 & 18 \\
UKP & ds70b & nuclear & 3 & 2000 & 79.3 & 75.9 & 78.2 & 77.0 & 0 & 0 \\
UKP & ds70b & nuclear & 4 & 2000 & 78.0 & 74.8 & 75.4 & 75.1 & 0 & 6 \\
UKP & ds70b & school & 1 & 2000 & 83.6 & 82.0 & 81.3 & 81.6 & 0 & 0 \\
UKP & ds70b & school & 2 & 2000 & 79.5 & 76.9 & 84.4 & 79.0 & 1 & 0 \\
UKP & ds70b & school & 3 & 2000 & 84.3 & 81.7 & 84.0 & 82.8 & 0 & 0 \\
UKP & ds70b & school & 4 & 2000 & 83.5 & 81.5 & 82.4 & 81.9 & 1 & 3 \\
UKP & ds70b & wage & 1 & 2000 & 78.6 & 76.7 & 76.9 & 76.8 & 0 & 0 \\
UKP & ds70b & wage & 2 & 2000 & 72.4 & 71.3 & 77.3 & 72.3 & 0 & 1 \\
UKP & ds70b & wage & 3 & 2000 & 78.8 & 76.4 & 79.4 & 77.6 & 0 & 0 \\
UKP & ds70b & wage & 4 & 2000 & 76.3 & 73.9 & 75.7 & 74.6 & 0 & 3 \\
UKP & oss20b & abortion & 1 & 3929 & 75.8 & 71.8 & 66.5 & 68.6 & 0 & 185 \\
UKP & oss20b & abortion & 2 & 3929 & 67.9 & 64.2 & 72.2 & 65.6 & 0 & 1108 \\
UKP & oss20b & abortion & 3 & 3929 & 76.8 & 75.0 & 66.2 & 69.4 & 0 & 15 \\
UKP & oss20b & abortion & 4 & 3929 & 74.8 & 68.7 & 67.8 & 68.2 & 0 & 540 \\
UKP & oss20b & cloning & 1 & 3039 & 76.5 & 78.2 & 73.4 & 75.0 & 0 & 116 \\
UKP & oss20b & cloning & 2 & 3039 & 73.3 & 73.1 & 77.5 & 73.5 & 0 & 760 \\
UKP & oss20b & cloning & 3 & 3039 & 75.5 & 79.9 & 70.3 & 73.1 & 0 & 8 \\
UKP & oss20b & cloning & 4 & 3039 & 76.4 & 76.6 & 74.3 & 75.2 & 0 & 257 \\
UKP & oss20b & death & 1 & 3651 & 78.6 & 79.2 & 69.6 & 73.0 & 0 & 186 \\
UKP & oss20b & death & 2 & 3651 & 71.8 & 68.4 & 77.3 & 69.9 & 0 & 1011 \\
UKP & oss20b & death & 3 & 3651 & 77.5 & 79.9 & 67.4 & 71.4 & 0 & 17 \\
UKP & oss20b & death & 4 & 3651 & 78.1 & 75.2 & 71.3 & 73.0 & 0 & 510 \\
UKP & oss20b & gun & 1 & 3341 & 74.7 & 71.5 & 67.2 & 68.8 & 0 & 157 \\
UKP & oss20b & gun & 2 & 3341 & 66.9 & 65.4 & 70.9 & 65.7 & 0 & 989 \\
UKP & oss20b & gun & 3 & 3341 & 75.0 & 75.2 & 66.1 & 68.9 & 0 & 15 \\
UKP & oss20b & gun & 4 & 3341 & 73.9 & 69.5 & 69.0 & 69.2 & 0 & 415 \\
UKP & oss20b & marijuana & 1 & 2475 & 74.2 & 76.6 & 69.3 & 71.8 & 0 & 125 \\
UKP & oss20b & marijuana & 2 & 2475 & 72.1 & 71.6 & 76.1 & 72.3 & 1 & 499 \\
UKP & oss20b & marijuana & 3 & 2475 & 73.2 & 79.0 & 66.4 & 69.8 & 0 & 8 \\
UKP & oss20b & marijuana & 4 & 2475 & 74.5 & 73.5 & 71.3 & 72.2 & 0 & 417 \\
UKP & oss20b & nuclear & 1 & 3576 & 79.8 & 77.8 & 75.1 & 76.3 & 0 & 139 \\
UKP & oss20b & nuclear & 2 & 3576 & 71.0 & 69.1 & 78.1 & 70.4 & 0 & 797 \\
UKP & oss20b & nuclear & 3 & 3576 & 79.7 & 79.3 & 73.2 & 75.7 & 0 & 10 \\
UKP & oss20b & nuclear & 4 & 3576 & 75.2 & 69.5 & 71.2 & 70.2 & 0 & 513 \\
UKP & oss20b & school & 1 & 3008 & 84.7 & 85.0 & 80.4 & 82.4 & 0 & 89 \\
UKP & oss20b & school & 2 & 3008 & 80.3 & 77.3 & 84.2 & 79.4 & 0 & 671 \\
UKP & oss20b & school & 3 & 3008 & 83.3 & 85.8 & 76.7 & 80.2 & 0 & 7 \\
UKP & oss20b & school & 4 & 3008 & 83.2 & 80.9 & 79.7 & 80.3 & 0 & 295 \\
UKP & oss20b & wage & 1 & 2473 & 80.4 & 79.1 & 76.6 & 77.5 & 0 & 142 \\
UKP & oss20b & wage & 2 & 2473 & 73.1 & 72.0 & 78.0 & 72.8 & 0 & 625 \\
UKP & oss20b & wage & 3 & 2473 & 79.5 & 79.6 & 74.7 & 76.3 & 0 & 9 \\
UKP & oss20b & wage & 4 & 2473 & 80.1 & 77.7 & 77.4 & 77.3 & 0 & 282 \\
UKP & oss120b & abortion & 1 & 3929 & 78.4 & 75.4 & 70.5 & 72.5 & 0 & 6 \\
UKP & oss120b & abortion & 2 & 3929 & 71.1 & 67.9 & 76.6 & 69.2 & 0 & 0 \\
UKP & oss120b & abortion & 3 & 3929 & 79.4 & 75.6 & 73.7 & 74.5 & 0 & 0 \\
UKP & oss120b & abortion & 4 & 3929 & 78.0 & 73.0 & 74.1 & 73.5 & 0 & 5 \\
UKP & oss120b & cloning & 1 & 3039 & 78.9 & 80.0 & 76.4 & 77.7 & 0 & 1 \\
UKP & oss120b & cloning & 2 & 3039 & 76.2 & 76.0 & 80.4 & 76.4 & 0 & 1 \\
UKP & oss120b & cloning & 3 & 3039 & 79.4 & 79.7 & 77.5 & 78.3 & 0 & 0 \\
UKP & oss120b & cloning & 4 & 3039 & 77.7 & 77.1 & 77.7 & 77.1 & 0 & 2 \\
UKP & oss120b & death & 1 & 3651 & 79.8 & 80.5 & 72.1 & 75.3 & 0 & 2 \\
UKP & oss120b & death & 2 & 3651 & 74.6 & 71.2 & 80.4 & 73.0 & 0 & 1 \\
UKP & oss120b & death & 3 & 3651 & 79.6 & 79.9 & 72.3 & 75.3 & 0 & 0 \\
UKP & oss120b & death & 4 & 3651 & 80.0 & 78.1 & 75.1 & 76.4 & 0 & 4 \\
UKP & oss120b & gun & 1 & 3341 & 77.2 & 75.4 & 71.2 & 72.9 & 0 & 3 \\
UKP & oss120b & gun & 2 & 3341 & 71.1 & 69.6 & 75.9 & 70.1 & 0 & 0 \\
UKP & oss120b & gun & 3 & 3341 & 78.4 & 76.5 & 73.7 & 74.7 & 0 & 0 \\
UKP & oss120b & gun & 4 & 3341 & 77.6 & 74.3 & 74.6 & 74.3 & 0 & 2 \\
UKP & oss120b & marijuana & 1 & 2475 & 77.4 & 78.8 & 73.9 & 75.8 & 0 & 0 \\
UKP & oss120b & marijuana & 2 & 2475 & 75.1 & 75.1 & 79.6 & 75.4 & 0 & 0 \\
UKP & oss120b & marijuana & 3 & 2475 & 79.6 & 80.6 & 76.5 & 78.1 & 0 & 0 \\
UKP & oss120b & marijuana & 4 & 2475 & 79.7 & 78.9 & 78.9 & 78.9 & 0 & 2 \\
UKP & oss120b & nuclear & 1 & 3576 & 81.1 & 79.8 & 76.1 & 77.7 & 0 & 3 \\
UKP & oss120b & nuclear & 2 & 3576 & 74.3 & 71.8 & 80.9 & 73.5 & 0 & 1 \\
UKP & oss120b & nuclear & 3 & 3576 & 59.2 & 46.2 & 33.5 & 25.3 & 0 & 0 \\
UKP & oss120b & nuclear & 4 & 3576 & 80.9 & 77.5 & 79.4 & 78.4 & 0 & 7 \\
UKP & oss120b & school & 1 & 3008 & 85.9 & 85.6 & 82.5 & 83.9 & 0 & 2 \\
UKP & oss120b & school & 2 & 3008 & 81.7 & 78.6 & 86.2 & 80.8 & 0 & 0 \\
UKP & oss120b & school & 3 & 3008 & 58.1 & 56.2 & 34.7 & 27.8 & 0 & 0 \\
UKP & oss120b & school & 4 & 3008 & 86.0 & 83.6 & 85.3 & 84.3 & 0 & 3 \\
UKP & oss120b & wage & 1 & 2473 & 80.8 & 80.3 & 76.6 & 77.8 & 0 & 2 \\
UKP & oss120b & wage & 2 & 2473 & 75.7 & 74.2 & 80.2 & 75.5 & 0 & 0 \\
UKP & oss120b & wage & 3 & 2473 & 82.2 & 81.3 & 78.8 & 79.6 & 0 & 0 \\
UKP & oss120b & wage & 4 & 2473 & 81.5 & 79.8 & 79.5 & 79.4 & 0 & 1 \\
UKP & gpt-5.2 & abortion & 1 & 2000 & 79.0 & 74.4 & 81.7 & 76.9 & 0 & 2 \\
UKP & gpt-5.2 & abortion & 2 & 2000 & 70.6 & 68.5 & 78.9 & 69.7 & 0 & 0 \\
UKP & gpt-5.2 & abortion & 3 & 1998 & 78.0 & 73.5 & 81.5 & 76.0 & 4 & 2 \\
UKP & gpt-5.2 & abortion & 4 & 2000 & 77.6 & 73.2 & 81.6 & 75.7 & 0 & 6 \\
UKP & gpt-5.2 & cloning & 1 & 2000 & 78.0 & 77.0 & 80.4 & 77.9 & 0 & 2 \\
UKP & gpt-5.2 & cloning & 2 & 2000 & 73.5 & 74.2 & 78.5 & 73.7 & 0 & 0 \\
UKP & gpt-5.2 & cloning & 3 & 2000 & 78.3 & 77.3 & 80.9 & 78.1 & 0 & 2 \\
UKP & gpt-5.2 & cloning & 4 & 2000 & 76.2 & 75.6 & 79.4 & 76.2 & 0 & 0 \\
UKP & gpt-5.2 & death & 1 & 2000 & 78.6 & 75.3 & 81.0 & 77.0 & 0 & 0 \\
UKP & gpt-5.2 & death & 2 & 2000 & 71.4 & 70.2 & 79.9 & 70.4 & 0 & 0 \\
UKP & gpt-5.2 & death & 3 & 1999 & 78.8 & 75.3 & 82.0 & 77.3 & 2 & 2 \\
UKP & gpt-5.2 & death & 4 & 1999 & 76.5 & 73.6 & 81.1 & 75.3 & 2 & 0 \\
UKP & gpt-5.2 & gun & 1 & 1998 & 75.9 & 73.5 & 79.6 & 74.8 & 4 & 2 \\
UKP & gpt-5.2 & gun & 2 & 2000 & 70.2 & 70.3 & 77.5 & 69.9 & 0 & 0 \\
UKP & gpt-5.2 & gun & 3 & 1999 & 76.9 & 74.2 & 80.0 & 75.6 & 2 & 6 \\
UKP & gpt-5.2 & gun & 4 & 1999 & 75.4 & 73.4 & 80.1 & 74.5 & 2 & 2 \\
UKP & gpt-5.2 & marijuana & 1 & 2000 & 79.7 & 78.4 & 82.7 & 79.6 & 0 & 0 \\
UKP & gpt-5.2 & marijuana & 2 & 2000 & 74.4 & 75.1 & 80.5 & 74.7 & 0 & 0 \\
UKP & gpt-5.2 & marijuana & 3 & 1998 & 79.4 & 78.2 & 82.6 & 79.4 & 4 & 0 \\
UKP & gpt-5.2 & marijuana & 4 & 1999 & 78.9 & 78.1 & 83.0 & 79.0 & 2 & 2 \\
UKP & gpt-5.2 & nuclear & 1 & 1999 & 79.5 & 75.7 & 82.3 & 77.9 & 2 & 0 \\
UKP & gpt-5.2 & nuclear & 2 & 2000 & 73.1 & 71.4 & 80.9 & 72.4 & 0 & 0 \\
UKP & gpt-5.2 & nuclear & 3 & 1999 & 79.1 & 75.1 & 82.1 & 77.3 & 2 & 2 \\
UKP & gpt-5.2 & nuclear & 4 & 2000 & 78.0 & 74.4 & 82.0 & 76.6 & 0 & 0 \\
UKP & gpt-5.2 & school & 1 & 2000 & 86.3 & 83.3 & 88.7 & 85.3 & 0 & 0 \\
UKP & gpt-5.2 & school & 2 & 2000 & 80.7 & 77.9 & 85.8 & 79.9 & 0 & 0 \\
UKP & gpt-5.2 & school & 3 & 1999 & 86.2 & 83.1 & 87.9 & 85.0 & 2 & 2 \\
UKP & gpt-5.2 & school & 4 & 1999 & 84.4 & 81.1 & 87.6 & 83.3 & 2 & 0 \\
UKP & gpt-5.2 & wage & 1 & 1998 & 82.8 & 80.4 & 84.3 & 81.8 & 4 & 2 \\
UKP & gpt-5.2 & wage & 2 & 1999 & 74.2 & 73.4 & 79.8 & 74.1 & 2 & 0 \\
UKP & gpt-5.2 & wage & 3 & 1999 & 82.5 & 80.1 & 84.2 & 81.7 & 2 & 0 \\
UKP & gpt-5.2 & wage & 4 & 1996 & 81.1 & 78.7 & 83.5 & 80.3 & 8 & 2 \\
\bottomrule
\end{longtable}

\end{document}